\documentclass{article}
\usepackage{microtype}
\usepackage{graphicx}
\usepackage{subfigure}
\usepackage{booktabs} 

\usepackage{hyperref}

\usepackage{amsmath,amssymb,mathtools,amsthm}
\usepackage{algorithmic}
\usepackage{wrapfig}

\usepackage{xspace}
\DeclarePairedDelimiterX{\infdivx}[2]{(}{)}{%
  #1\;\delimsize|\delimsize|\;#2%
}
\newcommand{\kld}[2]{\ensuremath{D_{KL}\infdivx{#1}{#2}}\xspace}

\makeatletter
\newcommand{\printfnsymbol}[1]{%
  \textsuperscript{\@fnsymbol{#1}}%
}
\makeatother

\usepackage[accepted]{icml2021}

\icmltitlerunning{Improved Denoising Diffusion Probabilistic Models}

\begin{document}

\twocolumn[
\icmltitle{Improved Denoising Diffusion Probabilistic Models}



\icmlsetsymbol{equal}{*}

\begin{icmlauthorlist}
\icmlauthor{Alex Nichol}{equal,openai}
\icmlauthor{Prafulla Dhariwal}{equal,openai}
\end{icmlauthorlist}
\icmlaffiliation{openai}{OpenAI, San Francisco, USA}
\icmlcorrespondingauthor{}{alex@openai.com}
\icmlcorrespondingauthor{}{prafulla@openai.com}

\icmlkeywords{Machine Learning, ICML, Neural Networks, Generative Models, Likelihood, Denoising Diffusion, Image Generation}

\vskip 0.3in
]



\printAffiliationsAndNotice{\icmlEqualContribution} 

\begin{abstract}
Denoising diffusion probabilistic models (DDPM) are a class of generative models which have recently been shown to produce excellent samples. We show that with a few simple modifications, DDPMs can also achieve competitive log-likelihoods while maintaining high sample quality. Additionally, we find that learning variances of the reverse diffusion process allows sampling with an order of magnitude fewer forward passes with a negligible difference in sample quality, which is important for the practical deployment of these models. We additionally use precision and recall to compare how well DDPMs and GANs cover the target distribution. Finally, we show that the sample quality and likelihood of these models scale smoothly with model capacity and training compute, making them easily scalable. We release our code at {\small \url{https://github.com/openai/improved-diffusion}}.
\end{abstract}

\section{Introduction}

\citet{diffusion} introduced diffusion probabilistic models, a class of generative models which match a data distribution by learning to reverse a gradual, multi-step noising process. More recently, \citet{ddpm} showed an equivalence between denoising diffusion probabilistic models (DDPM) and score based generative models \citep{scorematching,improvedscore}, which learn a gradient of the log-density of the data distribution using denoising score matching \citep{hyverianscorematching}. It has recently been shown that this class of models can produce high-quality images \citep{ddpm,improvedscore,adversarial} and audio \citep{wavegrad,diffwave}, but it has yet to be shown that DDPMs can achieve log-likelihoods competitive with other likelihood-based models such as autoregressive models \citep{pixelcnn} and VAEs \citep{vae}. This raises various questions, such as whether DDPMs are capable of capturing all the modes of a distribution. Furthermore, while \citet{ddpm} showed extremely good results on the CIFAR-10 \citep{cifar10} and LSUN \citep{lsun} datasets, it is unclear how well DDPMs scale to datasets with higher diversity such as ImageNet. Finally, while \citet{wavegrad} found that DDPMs can efficiently generate audio using a small number of sampling steps, it has yet to be shown that the same is true for images.

In this paper, we show that DDPMs can achieve log-likelihoods competitive with other likelihood-based models, even on high-diversity datasets like ImageNet. To more tightly optimise the variational lower-bound (VLB), we learn the reverse process variances using a simple reparameterization and a hybrid learning objective that combines the VLB with the simplified objective from \citet{ddpm}.

We find surprisingly that, with our hybrid objective, our models obtain better log-likelihoods than those obtained by optimizing the log-likelihood directly, and discover that the latter objective has much more gradient noise during training. We show that a simple importance sampling technique reduces this noise and allows us to achieve better log-likelihoods than with the hybrid objective.

After incorporating learned variances into our model, we surprisingly discovered that we could sample in fewer steps from our models with very little change in sample quality. While DDPM \citep{ddpm} requires hundreds of forward passes to produce good samples, we can achieve good samples with as few as 50 forward passes, thus speeding up sampling for use in practical applications. In parallel to our work, \citet{ddim} develops a different approach to fast sampling, and we compare against their approach, DDIM, in our experiments.

While likelihood is a good metric to compare against other likelihood-based models, we also wanted to compare the distribution coverage of these models with GANs. We use the improved precision and recall metrics \citep{improvedpr} and discover that diffusion models achieve much higher recall for similar FID, suggesting that they do indeed cover a much larger portion of the target distribution. 

Finally, since we expect machine learning models to consume more computational resources in the future, we evaluate the performance of these models as we increase model size and training compute. Similar to \cite{scalingcompendium}, we observe trends that suggest predictable improvements in performance as we increase training compute. 

\section{Denoising Diffusion Probabilistic Models}
\label{sec:ddpm}
We briefly review the formulation of DDPMs from \citet{ddpm}. This formulation makes various simplifying assumptions, such as a fixed noising process $q$ which adds diagonal Gaussian noise at each timestep. For a more general derivation, see \citet{diffusion}.

\subsection{Definitions}

Given a data distribution $x_0 \sim q(x_0)$, we define a forward noising process $q$ which produces latents $x_1$ through $x_T$ by adding Gaussian noise at time $t$ with variance $\beta_t \in (0,1)$ as follows:
\begin{alignat}{2}
    q(x_1, ..., x_T | x_0) &\coloneqq \prod_{t=1}^{T} q(x_t | x_{t-1}) \label{eq:joint} \\
    q(x_t | x_{t-1}) &\coloneqq \mathcal{N}(x_t; \sqrt{1-\beta_t} x_{t-1}, \beta_t \mathbf{I}) \label{eq:singlestep}
\end{alignat}

Given sufficiently large $T$ and a well behaved schedule of $\beta_t$, the latent $x_T$ is nearly an isotropic Gaussian distribution. Thus, if we know the exact reverse distribution $q(x_{t-1}|x_t)$, we can sample $x_T \sim \mathcal{N}(0, \mathbf{I})$ and run the process in reverse to get a sample from $q(x_0)$. However, since $q(x_{t-1}|x_t)$ depends on the entire data distribution, we approximate it using a neural network as follows:
\begin{equation}
\label{eq:nn}
p_{\theta}(x_{t-1}|x_t) \coloneqq \mathcal{N}(x_{t-1}; \mu_{\theta}(x_t, t), \Sigma_{\theta}(x_t, t))
\end{equation}

The combination of $q$ and $p$ is a variational auto-encoder \citep{vae}, and we can write the variational lower bound (VLB) as follows:
\begin{alignat}{2}
    L_{\text{vlb}} &\coloneqq L_0 + L_1 + ... + L_{T-1} + L_T \label{eq:loss} \\
    L_{0} &\coloneqq -\log p_{\theta}(x_0 | x_1) \label{eq:loss0} \\
    L_{t-1} &\coloneqq \kld{q(x_{t-1}|x_t,x_0)}{p_{\theta}(x_{t-1}|x_t)} \label{eq:losst} \\
    L_{T} &\coloneqq \kld{q(x_T | x_0)}{p(x_T)} \label{eq:lossT} 
\end{alignat}

Aside from $L_0$, each term of Equation \ref{eq:loss} is a \ensuremath{KL} divergence between two Gaussians, and can thus be evaluated in closed form. To evaluate $L_0$ for images, we assume that each color component is divided into 256 bins, and we compute the probability of $p_{\theta}(x_0 | x_1)$ landing in the correct bin (which is tractable using the CDF of the Gaussian distribution). Also note that while $L_T$ does not depend on $\theta$, it will be close to zero if the forward noising process adequately destroys the data distribution so that $q(x_T|x_0) \approx \mathcal{N}(0, \mathbf{I})$.

As noted in \cite{ddpm}, the noising process defined in Equation \ref{eq:singlestep} allows us to sample an arbitrary step of the noised latents directly conditioned on the input $x_0$. With $\alpha_t \coloneqq 1 - \beta_t$ and $\bar{\alpha}_t \coloneqq \prod_{s=0}^{t} \alpha_s$, we can write the marginal
\begin{alignat}{2}
    q(x_t|x_0) &= \mathcal{N}(x_t; \sqrt{\bar{\alpha}_t} x_0, (1-\bar{\alpha}_t) \mathbf{I}) \label{eq:marginal} \\
    x_t &=  \sqrt{\bar{\alpha}_t} x_0 + \sqrt{1-\bar{\alpha}_t} \epsilon \label{eq:jumpnoise}
\end{alignat}
where $\epsilon \sim \mathcal{N}(0,\mathbf{I})$. Here, $1 - \bar{\alpha}_t$ tells us the variance of the noise for an arbitrary timestep, and we could equivalently use this to define the noise schedule instead of $\beta_t$.

Using Bayes theorem, one can calculate the posterior $q(x_{t-1}|x_t,x_0)$ in terms of $\tilde{\beta}_t$ and $\tilde{\mu}_t(x_t,x_0)$ which are defined as follows:
\begin{alignat}{2}
    \tilde{\beta}_t &\coloneqq \frac{1-\bar{\alpha}_{t-1}}{1-\bar{\alpha}_t} \beta_t \label{eq:betatilde} \\
    \tilde{\mu}_t(x_t,x_0) &\coloneqq \frac{\sqrt{\bar{\alpha}_{t-1}}\beta_t}{1-\bar{\alpha}_t}x_0 + \frac{\sqrt{\alpha_t}(1-\bar{\alpha}_{t-1})}{1-\bar{\alpha}_t} x_t \label{eq:mutilde} \\
    q(x_{t-1}|x_t,x_0) &= \mathcal{N}(x_{t-1}; \tilde{\mu}(x_t, x_0), \tilde{\beta}_t \mathbf{I}) \label{eq:posterior}
\end{alignat}

\subsection{Training in Practice}

The objective in Equation \ref{eq:loss} is a sum of independent terms $L_{t-1}$, and Equation \ref{eq:jumpnoise} provides an efficient way to sample from an arbitrary step of the forward noising process and estimate $L_{t-1}$ using the posterior (Equation \ref{eq:posterior}) and prior (Equation \ref{eq:nn}). We can thus randomly sample $t$ and use the expectation $E_{t,x_0,\epsilon}[L_{t-1}]$ to estimate $L_{\text{vlb}}$. \citet{ddpm} uniformly sample $t$ for each image in each mini-batch.

There are many different ways to parameterize $\mu_{\theta}(x_t, t)$ in the prior. The most obvious option is to predict $\mu_{\theta}(x_t, t)$ directly with a neural network. Alternatively, the network could predict $x_0$, and this output could be used in Equation \ref{eq:mutilde} to produce $\mu_{\theta}(x_t, t)$. The network could also predict the noise $\epsilon$ and use Equations \ref{eq:jumpnoise} and \ref{eq:mutilde} to derive
\begin{equation}
\mu_{\theta}(x_t, t) = \frac{1}{\sqrt{\alpha_t}} \left( x_t - \frac{\beta_t}{\sqrt{1-\bar{\alpha}_t}} \epsilon_{\theta}(x_t, t) \right)
\end{equation}
\citet{ddpm} found that predicting $\epsilon$ worked best, especially when combined with a reweighted loss function:
\begin{equation}
L_{\text{simple}} = E_{t,x_0,\epsilon}\left[ || \epsilon - \epsilon_{\theta}(x_t, t) ||^2 \right]
\end{equation}
This objective can be seen as a reweighted form of $L_{\text{vlb}}$ (without the terms affecting $\Sigma_{\theta}$). The authors found that optimizing this reweighted objective resulted in much better sample quality than optimizing $L_{\text{vlb}}$ directly, and explain this by drawing a connection to generative score matching \citep{scorematching,improvedscore}.

One subtlety is that $L_{\text{simple}}$ provides no learning signal for $\Sigma_{\theta}(x_t, t)$. This is irrelevant, however, since \citet{ddpm} achieved their best results by fixing the variance to $\sigma_t^2 \mathbf{I}$ rather than learning it. They found that they achieve similar sample quality using either $\sigma_t^2 = \beta_t$ or $\sigma_t^2 = \tilde{\beta}_t$, which are the upper and lower bounds on the variance given by $q(x_0)$ being either isotropic Gaussian noise or a delta function, respectively.

\section{Improving the Log-likelihood}
\label{sec:improvinglikelihood}

While \citet{ddpm} found that DDPMs can generate high-fidelity samples according to FID \citep{fid} and Inception Score \citep{inceptionscore}, they were unable to achieve competitive log-likelihoods with these models. Log-likelihood is a widely used metric in generative modeling, and it is generally believed that optimizing log-likelihood forces generative models to capture all of the modes of the data distribution \citep{vqvae2}. Additionally, recent work \citep{scalingcompendium} has shown that small improvements in log-likelihood can have a dramatic impact on sample quality and learnt feature representations. Thus, it is important to explore why DDPMs seem to perform poorly on this metric, since this may suggest a fundamental shortcoming such as bad mode coverage. This section explores several modifications to the algorithm described in Section \ref{sec:ddpm} that, when combined, allow DDPMs to achieve much better log-likelihoods on image datasets, suggesting that these models enjoy the same benefits as other likelihood-based generative models.

To study the effects of different modifications, we train fixed model architectures with fixed hyperparameters on the ImageNet $64 \times 64$ \citep{imagenet64} and CIFAR-10 \citep{cifar10} datasets. While CIFAR-10 has seen more usage for this class of models, we chose to study ImageNet $64 \times 64$ as well because it provides a good trade-off between diversity and resolution, allowing us to train models quickly without worrying about overfitting. Additionally, ImageNet $64 \times 64$ has been studied extensively in the context of generative modeling \citep{pixelcnn,spn,sparsetransformer,routingtransformer}, allowing us to compare DDPMs directly to many other generative models.

The setup from \citet{ddpm} (optimizing $L_{\text{simple}}$ while setting $\sigma_t^2 = \beta_t$ and $T = 1000$) achieves a log-likelihood of 3.99 (bits/dim) on ImageNet $64 \times 64$ after 200K training iterations. We found in early experiments that we could get a boost in log-likelihood by increasing $T$ from 1000 to 4000; with this change, the log-likelihood improves to 3.77. For the remainder of this section, we use $T = 4000$, but we explore this choice in Section \ref{sec:numberofsteps}.

\subsection{Learning $\Sigma_{\theta}(x_t, t)$}
\label{sec:learnsigma}

\begin{figure}[ht]
\begin{center}
\centerline{\includegraphics[width=0.8\columnwidth]{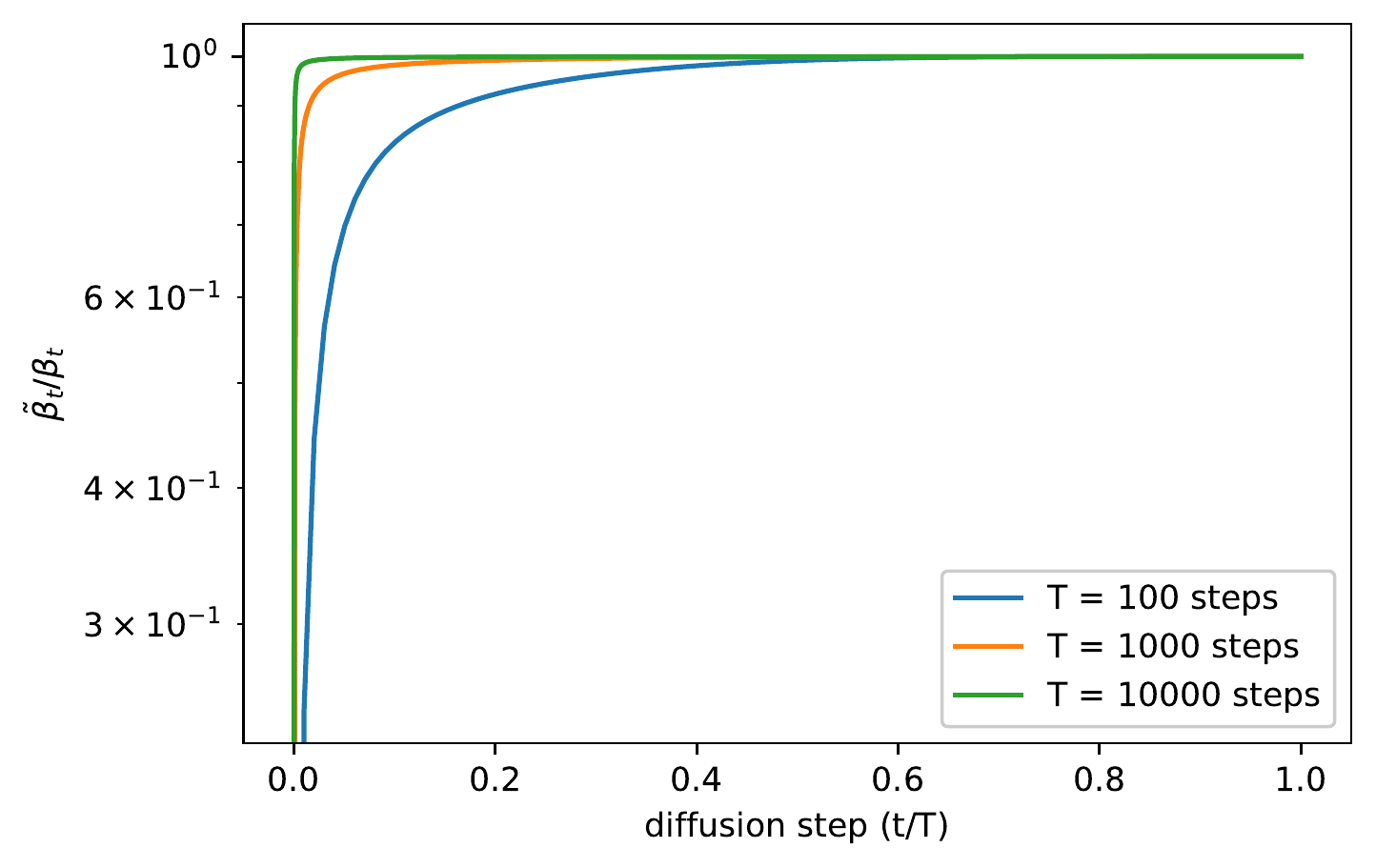}}
\caption{\label{fig:sigmaratio} The ratio $\tilde{\beta}_t / \beta_t$ for every diffusion step for diffusion processes of different lengths.}
\end{center}
\vskip -0.4in
\end{figure}

\begin{figure}[ht]
\vskip 0.2in
\begin{center}
\centerline{\includegraphics[width=0.8\columnwidth]{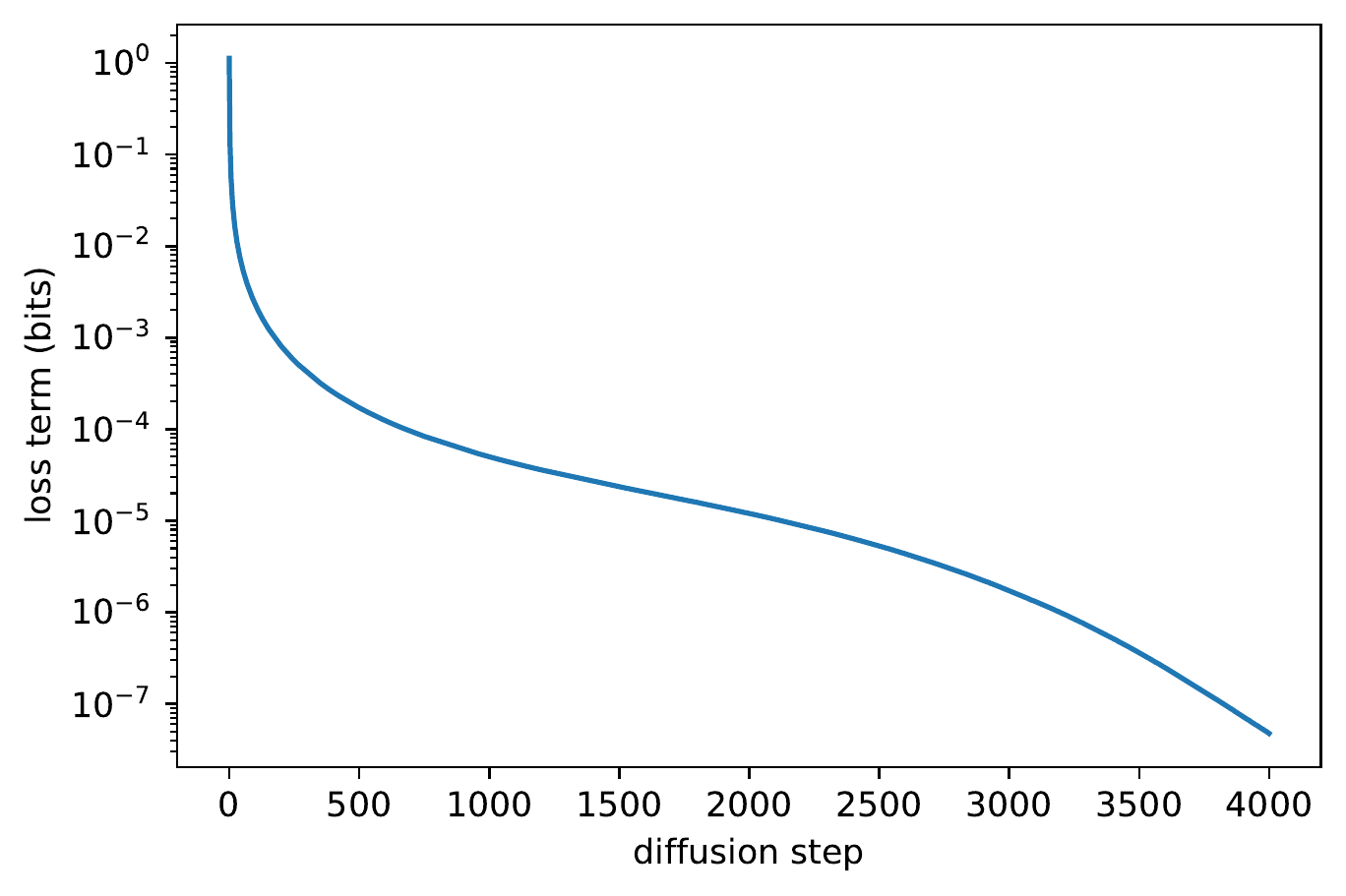}}
\caption{\label{fig:lossterms} Terms of the VLB vs diffusion step. The first few terms contribute most to NLL.}\end{center}
\vskip -0.4in
\end{figure}


In \citet{ddpm}, the authors set $\Sigma_{\theta}(x_t, t) = \sigma^2_t \mathbf{I}$, where $\sigma_t$ is not learned. Oddly, they found that fixing $\sigma^2_t$ to $\beta_t$ yielded roughly the same sample quality as fixing it to $\tilde{\beta}_t$. Considering that $\beta_t$ and $\tilde{\beta}_t$ represent two opposite extremes, it is reasonable to ask why this choice doesn't affect samples. One clue is given by Figure \ref{fig:sigmaratio}, which shows that $\beta_t$ and $\tilde{\beta}_t$ are almost equal except near $t = 0$, i.e. where the model is dealing with imperceptible details. Furthermore, as we increase the number of diffusion steps, $\beta_t$ and $\tilde{\beta}_t$ seem to remain close to one another for more of the diffusion process. This suggests that, in the limit of infinite diffusion steps, the choice of $\sigma_t$ might not matter at all for sample quality. In other words, as we add more diffusion steps, the model mean $\mu_{\theta}(x_t, t)$ determines the distribution much more than $\Sigma_{\theta}(x_t, t)$.

While the above argument suggests that fixing $\sigma_t$ is a reasonable choice for the sake of sample quality, it says nothing about log-likelihood. In fact, Figure \ref{fig:lossterms} shows that the first few steps of the diffusion process contribute the most to the variational lower bound. Thus, it seems likely that we could improve log-likelihood by using a better choice of $\Sigma_{\theta}(x_t, t)$. To achieve this, we must learn $\Sigma_{\theta}(x_t, t)$ without the instabilities encountered by \citet{ddpm}.

Since Figure \ref{fig:sigmaratio} shows that the reasonable range for $\Sigma_{\theta}(x_t, t)$ is very small, it would be hard for a neural network to predict $\Sigma_{\theta}(x_t, t)$ directly, even in the log domain, as observed by \citet{ddpm}. Instead, we found it better to parameterize the variance as an interpolation between $\beta_t$ and $\tilde{\beta}_t$ in the log domain. In particular, our model outputs a vector $v$ containing one component per dimension, and we turn this output into variances as follows:
\begin{equation}
\Sigma_{\theta}(x_t, t) = \exp(v \log \beta_t + (1-v) \log \tilde{\beta}_t)
\end{equation}

We did not apply any constraints on $v$, theoretically allowing the model to predict variances outside of the interpolated range. However, we did not observe the network doing this in practice, suggesting that the bounds for $\Sigma_{\theta}(x_t, t)$ are indeed expressive enough.

Since $L_{\text{simple}}$ doesn't depend on $\Sigma_{\theta}(x_t, t)$, we define a new hybrid objective:
\begin{equation}
L_{\text{hybrid}} = L_{\text{simple}} + \lambda L_{\text{vlb}}
\end{equation}

For our experiments, we set $\lambda = 0.001$ to prevent $L_{\text{vlb}}$ from overwhelming $L_{\text{simple}}$. Along this same line of reasoning, we also apply a stop-gradient to the $\mu_{\theta}(x_t, t)$ output for the $L_{\text{vlb}}$ term. This way, $L_{\text{vlb}}$ can guide $\Sigma_{\theta}(x_t, t)$ while $L_{\text{simple}}$ is still the main source of influence over $\mu_{\theta}(x_t, t)$.

\subsection{Improving the Noise Schedule}
\label{sec:schedule}

\begin{figure}[t]
\begin{center}
\centerline{\includegraphics[width=\columnwidth]{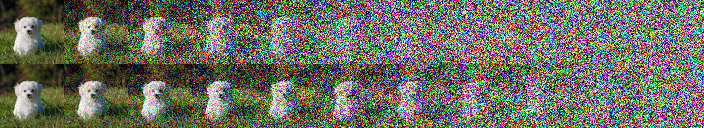}}
\caption{\label{fig:noisingprocess} Latent samples from linear (top) and cosine (bottom) schedules respectively at linearly spaced values of $t$ from $0$ to $T$. The latents in the last quarter of the linear schedule are almost purely noise, whereas the cosine schedule adds noise more slowly}
\end{center}
\vskip -0.2in
\end{figure}

\begin{figure}[t]
\begin{center}
\centerline{\includegraphics[width=0.8\columnwidth]{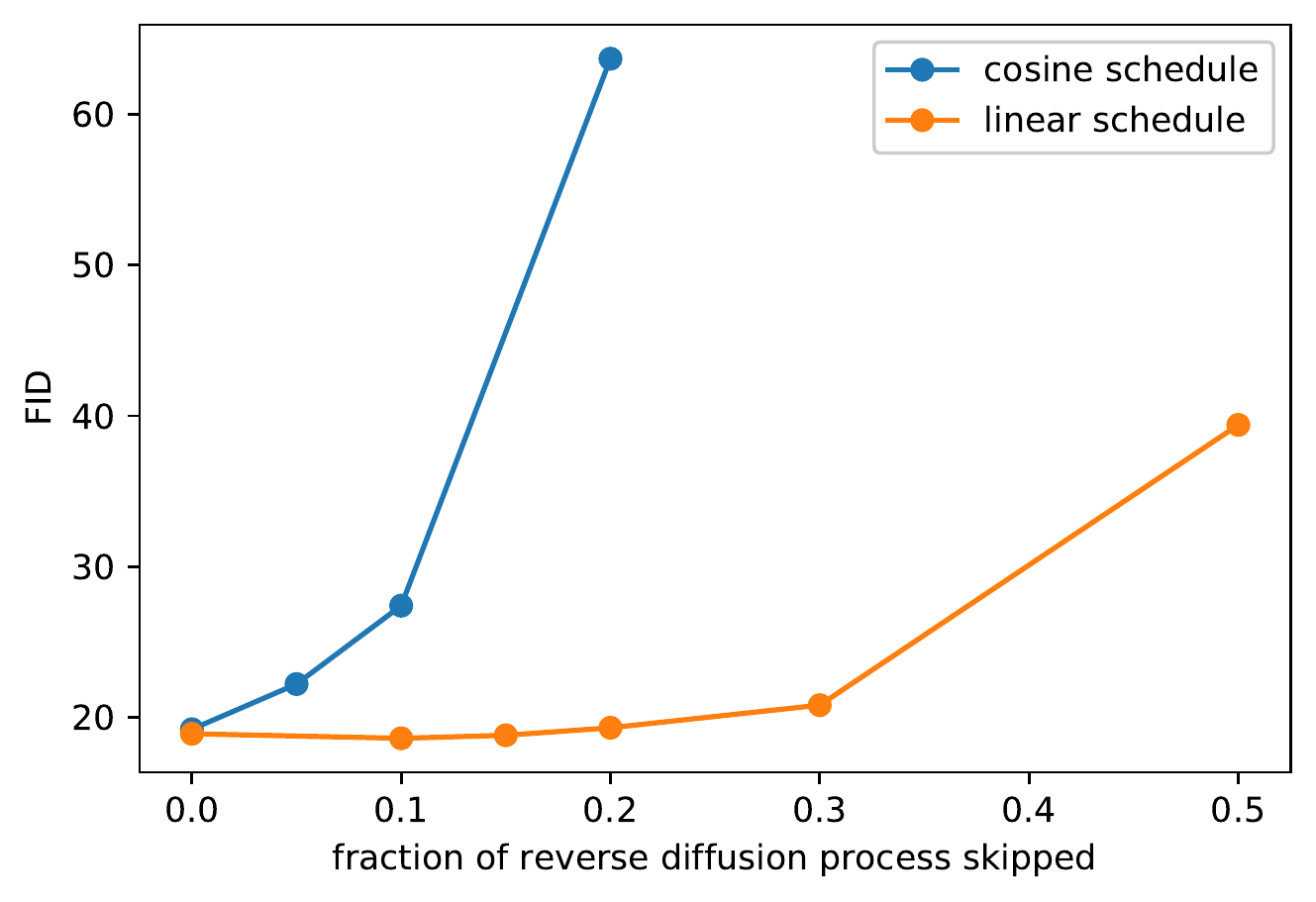}}
\caption{\label{fig:skipsteps} FID when skipping a prefix of the reverse diffusion process on ImageNet $64 \times 64$.}
\end{center}
\end{figure}

\begin{figure}[ht]
\begin{center}
\centerline{\includegraphics[width=0.8\columnwidth]{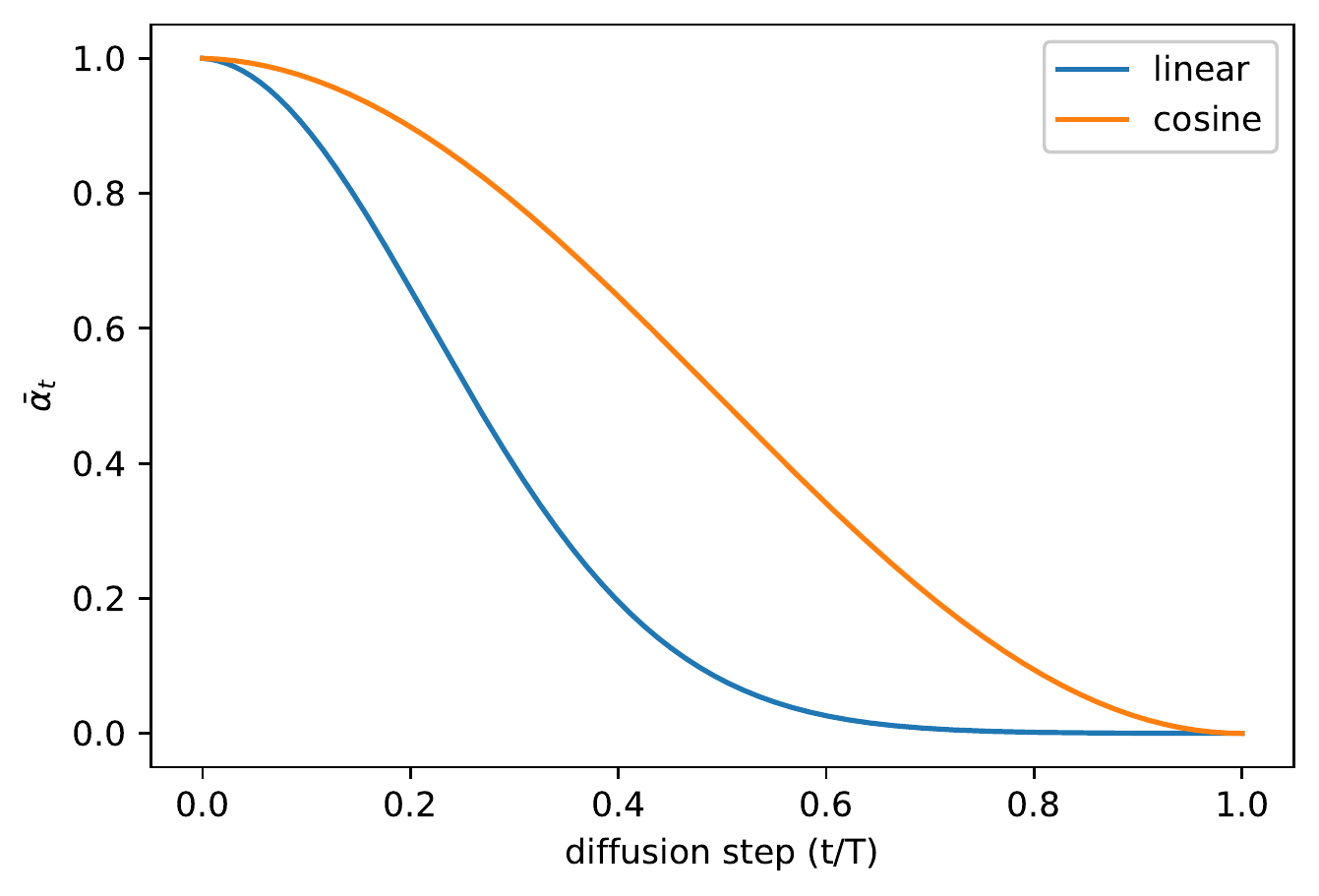}}
\caption{\label{fig:alphabar} $\bar{\alpha}_t$ throughout diffusion in the linear schedule and our proposed cosine schedule.}
\end{center}
\vskip -0.2in
\end{figure}


We found that while the linear noise schedule used in \citet{ddpm} worked well for high resolution images, it was sub-optimal for images of resolution $64 \times 64$ and $32 \times 32$. In particular, the end of the forward noising process is too noisy, and so doesn't contribute very much to sample quality. This can be seen visually in Figure \ref{fig:noisingprocess}. The result of this effect is studied in Figure \ref{fig:skipsteps}, where we see that a model trained with the linear schedule does not get much worse (as measured by FID) when we skip up to 20\% of the reverse diffusion process.

To address this problem, we construct a different noise schedule in terms of $\bar{\alpha}_t$:
\begin{align}
\bar{\alpha}_t = \frac{f(t)}{f(0)},\;\; f(t) = \cos\left(\frac{t/T + s}{1+s}\cdot\frac{\pi}{2}\right)^2
\end{align}
To go from this definition to variances $\beta_t$, we note that $
\beta_t = 1 - \frac{\bar{\alpha}_t}{\bar{\alpha}_{t-1}}$. In practice, we clip $\beta_t$ to be no larger than 0.999 to prevent singularities at the end of the diffusion process near $t=T$.

Our cosine schedule is designed to have a linear drop-off of $\bar{\alpha}_t$ in the middle of the process, while changing very little near the extremes of $t=0$ and $t=T$ to prevent abrupt changes in noise level. Figure \ref{fig:alphabar} shows how $\bar{\alpha}_t$ progresses for both schedules. We can see that the linear schedule from \citet{ddpm} falls towards zero much faster, destroying information more quickly than necessary.

We use a small offset $s$ to prevent $\beta_t$ from being too small near $t=0$, since we found that having tiny amounts of noise at the beginning of the process made it hard for the network to predict $\epsilon$ accurately enough. In particular, we selected $s$ such that $\sqrt{\beta_0}$ was slightly smaller than the pixel bin size $1/127.5$, which gives $s=0.008$. We chose to use $cos^2$ in particular because it is a common mathematical function with the shape we were looking for. This choice was arbitrary, and we expect that many other functions with similar shapes would work as well.

\subsection{Reducing Gradient Noise}
\label{sec:gradnoise}

\begin{figure}[ht]
\begin{center}
\centerline{\includegraphics[width=0.8\columnwidth]{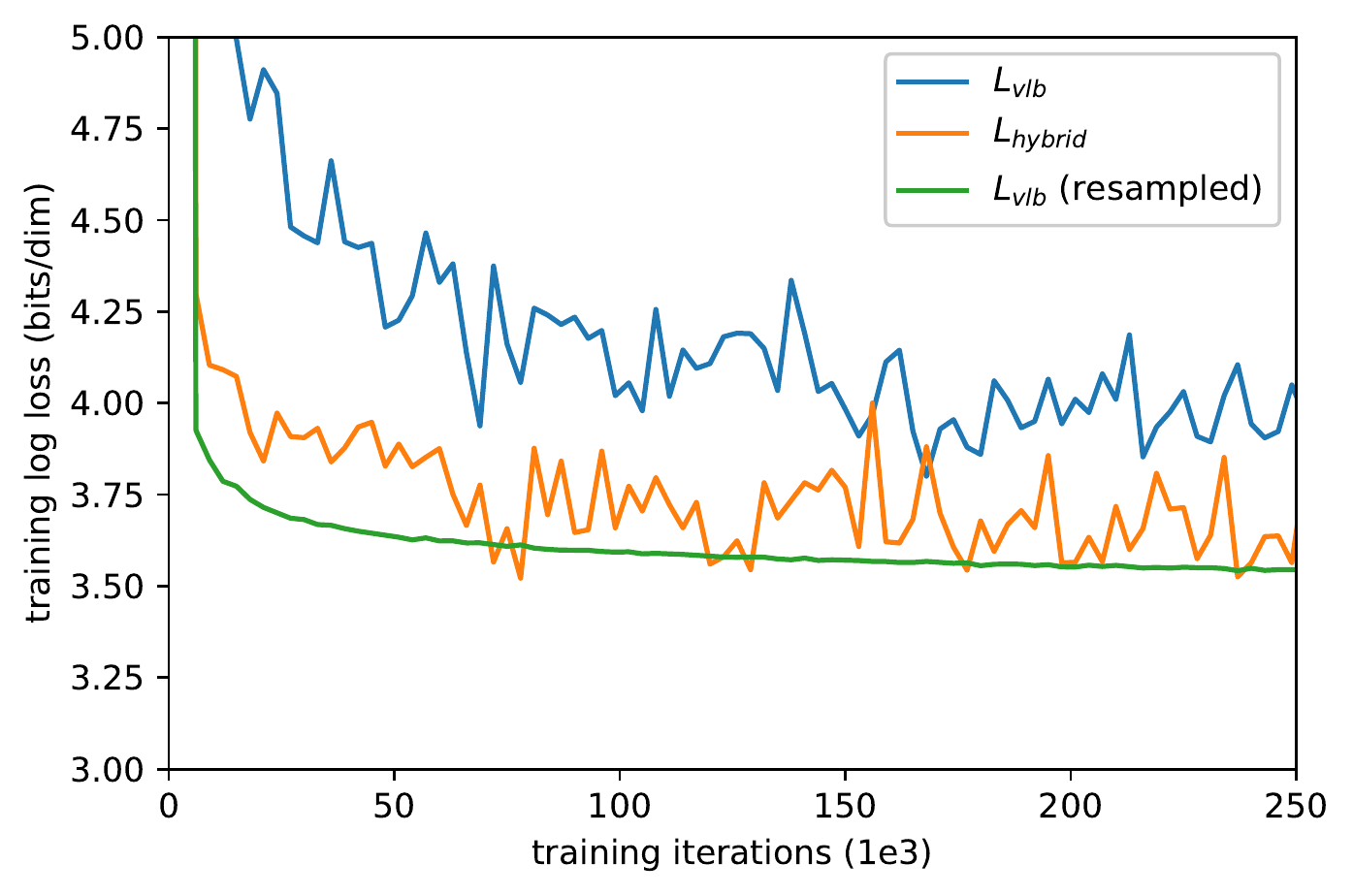}}
\caption{\label{fig:reweightlosscurves} Learning curves comparing the log-likelihoods achieved by different objectives on ImageNet $64 \times 64$.}
\end{center}
\vskip -0.4in
\end{figure}

\begin{figure}[ht]
\vskip 0.2in
\begin{center}
\centerline{\includegraphics[width=0.8\columnwidth]{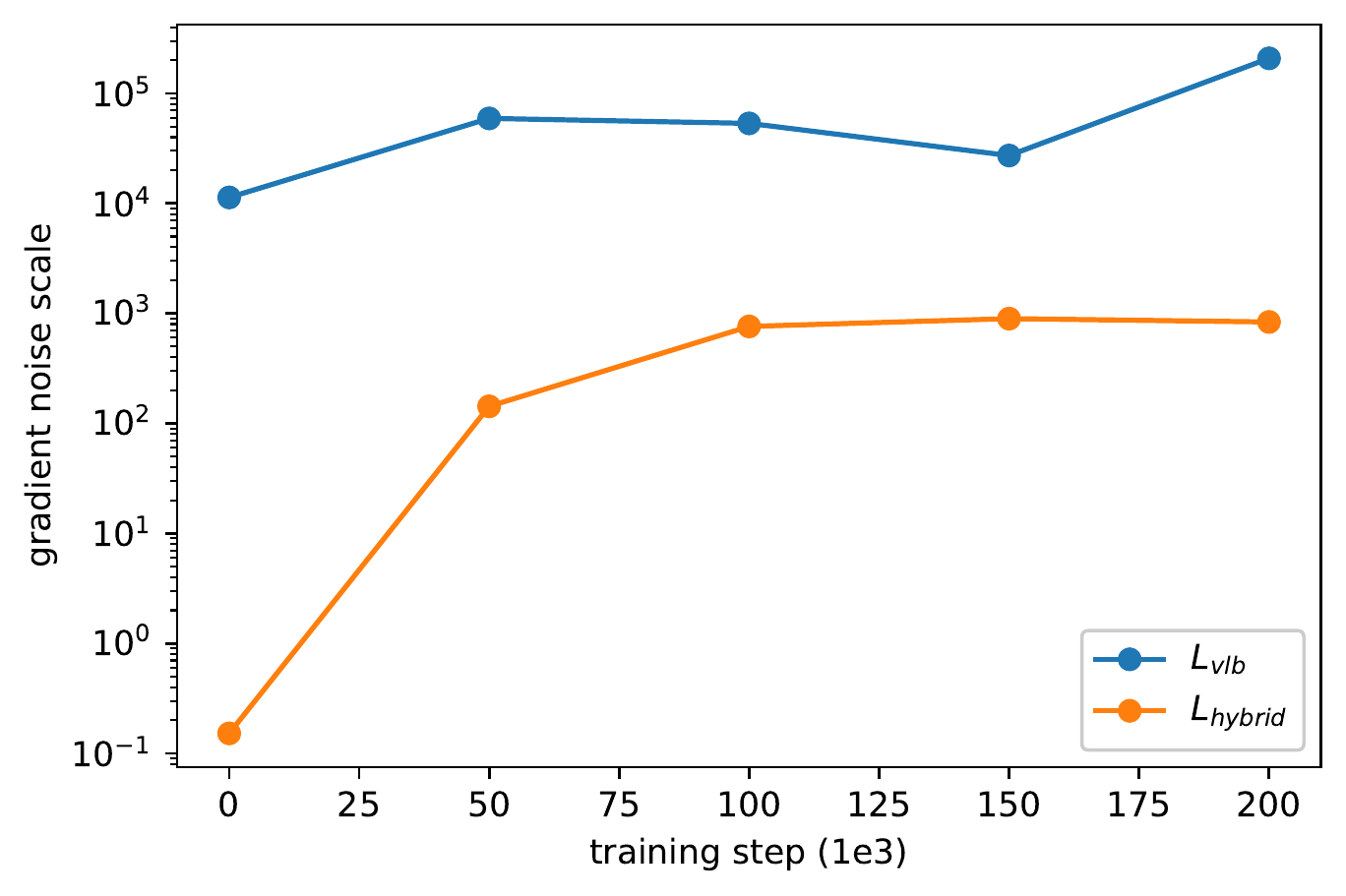}}
\caption{\label{fig:noisescales} Gradient noise scales for the $L_{\text{vlb}}$ and $L_{\text{hybrid}}$ objectives on ImageNet $64 \times 64$.}
\end{center}
\vskip -0.2in
\end{figure}

We expected to achieve the best log-likelihoods by optimizing $L_{\text{vlb}}$ directly, rather than by optimizing $L_{\text{hybrid}}$. However, we were surprised to find that $L_{\text{vlb}}$ was actually quite difficult to optimize in practice, at least on the diverse ImageNet $64 \times 64$ dataset. Figure \ref{fig:reweightlosscurves} shows the learning curves for both $L_{\text{vlb}}$ and $L_{\text{hybrid}}$. Both curves are noisy, but the hybrid objective clearly achieves better log-likelihoods on the training set given the same amount of training time.

We hypothesized that the gradient of $L_{\text{vlb}}$ was much noisier than that of $L_{\text{hybrid}}$. We confirmed this by evaluating the gradient noise scales \citep{gradnoisescale} for models trained with both objectives, as shown in Figure \ref{fig:noisescales}. Thus, we sought out a way to reduce the variance of $L_{\text{vlb}}$ in order to optimize directly for log-likelihood.

Noting that different terms of $L_{\text{vlb}}$ have greatly different magnitudes (Figure \ref{fig:lossterms}), we hypothesized that sampling $t$ uniformly causes unnecessary noise in the $L_{\text{vlb}}$ objective. To address this, we employ importance sampling:
\begin{equation}
L_{\text{vlb}} = E_{t \sim p_t}\left[\frac{L_t}{p_t}\right]\text{, where } p_t \propto \sqrt{E[L_t^2]}\text{ and }\sum p_t = 1
\end{equation}
Since $E[L_t^2]$ is unknown beforehand and may change throughout training, we maintain a history of the previous 10 values for each loss term, and update this dynamically during training. At the beginning of training, we sample $t$ uniformly until we draw 10 samples for every $t \in [0,T-1]$.

With this importance sampled objective, we are able to achieve our best log-likelihoods by optimizing $L_{\text{vlb}}$. This can be seen in Figure \ref{fig:reweightlosscurves} as the $L_{\text{vlb}}$ (resampled) curve. The figure also shows that the importance sampled objective is considerably less noisy than the original, uniformly sampled objective. We found that the importance sampling technique was not helpful when optimizing the less-noisy $L_{\text{hybrid}}$ objective directly.

\subsection{Results and Ablations}
\label{sec:summarizing}

In this section, we ablate the changes we have made to achieve better log-likelihoods. Table \ref{tbl:improvementcomparison} summarizes the results of our ablations on ImageNet $64 \times 64$, and Table \ref{tbl:cifar_improvementcomparison} shows them for CIFAR-10. We also trained our best ImageNet $64 \times 64$ models for 1.5M iterations, and report these results as well. $L_{\text{vlb}}$ and $L_{\text{hybrid}}$ were trained with learned sigmas using the parameterization from Section \ref{sec:learnsigma}. For $L_{\text{vlb}}$, we used the resampling scheme from Section \ref{sec:gradnoise}.

Based on our ablations, using $L_{\text{hybrid}}$ and our cosine schedule improves log-likelihood while keeping similar FID as the baseline from \citet{ddpm}. Optimizing $L_{\text{vlb}}$ further improves log-likelihood at the cost of a higher FID. We generally prefer to use $L_{\text{hybrid}}$ over $L_{\text{vlb}}$ as it gives a boost in likelihood without sacrificing sample quality. 

In Table \ref{tbl:allvlbcomparison} we compare our best likelihood models against prior work, showing that these models are competitive with the best conventional methods in terms of log-likelihood.

\begin{table}[t]
    \caption{\label{tbl:improvementcomparison} Ablating schedule and objective on ImageNet $64 \times 64$.}
    \centering
    \vskip 0.15in
	\begin{center}
	\begin{small}
    \begin{tabular}{ccccll}
    	\toprule
        Iters & $T$ & Schedule & Objective & NLL & FID \\ 
		\midrule
        200K & 1K & linear & $L_{\text{simple}}$ & 3.99 & 32.5 \\
        200K & 4K & linear & $L_{\text{simple}}$ & 3.77 & 31.3 \\
		\midrule
        200K & 4K & linear & $L_{\text{hybrid}}$ & 3.66 & 32.2 \\
        200K & 4K & cosine & $L_{\text{simple}}$ & 3.68 & \bf 27.0 \\
        200K & 4K & cosine & $L_{\text{hybrid}}$ & 3.62 & 28.0 \\
        200K & 4K & cosine & $L_{\text{vlb}}$ & \bf 3.57 & 56.7 \\
        \midrule
        1.5M & 4K & cosine & $L_{\text{hybrid}}$ & 3.57 & \bf 19.2 \\
        1.5M & 4K & cosine & $L_{\text{vlb}}$ & \bf 3.53 & 40.1 \\
        \bottomrule
    \end{tabular}
    \end{small}
    \end{center}
    \vskip -0.1in
\end{table}

\begin{table}[t]
    \caption{\label{tbl:cifar_improvementcomparison} Ablating schedule and objective on CIFAR-10.}
    \centering
    \vskip 0.15in
	\begin{center}
	\begin{small}
    \begin{tabular}{ccccll}
    	\toprule
        Iters & $T$ & Schedule & Objective & NLL & FID \\ 
        \midrule
        500K & 1K & linear & $L_{\text{simple}}$ & 3.73 & 3.29 \\
        500K & 4K & linear & $L_{\text{simple}}$ & 3.37 & \bf 2.90 \\
        \midrule
        500K & 4K & linear & $L_{\text{hybrid}}$ & 3.26 & 3.07 \\
        500K & 4K & cosine & $L_{\text{simple}}$ & 3.26 & 3.05 \\
        500K & 4K & cosine & $L_{\text{hybrid}}$ & 3.17 & 3.19 \\
        500K & 4K & cosine & $L_{\text{vlb}}$ & \bf 2.94 & 11.47 \\
        \bottomrule
    \end{tabular}
    \end{small}
    \end{center}
    \vskip -0.1in
\end{table}

\begin{table}[t]
    \caption{\label{tbl:allvlbcomparison} Comparison of DDPMs to other likelihood-based models on CIFAR-10 and Unconditional ImageNet $64 \times 64$. NLL is reported in bits/dim. On ImageNet $64 \times 64$, our model is competitive with the best convolutional models, but is worse than fully transformer-based architectures.}
    \centering
    \vskip 0.15in
	\begin{center}
	\begin{small}
    \begin{tabular}{lcc}
    	\toprule
        Model & ImageNet & CIFAR \\
        \midrule
        Glow \citep{glow} & 3.81 & 3.35 \\
        Flow++ \citep{flow++} & 3.69 & 3.08 \\
        PixelCNN \citep{pixelcnn} & 3.57 & 3.14 \\
        SPN \citep{spn} & 3.52 & - \\
        NVAE \citep{nvae} & - & 2.91 \\
        Very Deep VAE \citep{vdvae} & 3.52 & 2.87 \\
        PixelSNAIL \citep{pixelsnail} & 3.52 & 2.85 \\
        Image Transformer \citep{imagetransformer} & 3.48 & 2.90 \\
        Sparse Transformer \citep{sparsetransformer} & 3.44 & \bf 2.80 \\
        Routing Transformer \citep{routingtransformer} & \bf 3.43 & - \\
        \midrule 
        DDPM \citep{ddpm} & 3.77 & 3.70 \\
        DDPM (cont flow) \citep{sde} & - & 2.99 \\
        Improved DDPM (ours) & \bf 3.53 & \bf 2.94 \\
        \bottomrule
    \end{tabular}
    \end{small}
    \end{center}
    \vskip -0.1in
\end{table}

\section{Improving Sampling Speed}
\label{sec:numberofsteps}

All of our models were trained with 4000 diffusion steps, and thus producing a single sample takes several minutes on a modern GPU. In this section, we explore how performance scales if we reduce the steps used during sampling, and find that our pre-trained $L_{\text{hybrid}}$ models can produce high-quality samples with many fewer diffusion steps than they were trained with (without any fine-tuning). Reducing the steps in this way makes it possible to sample from our models in a number of seconds rather than minutes, and greatly improves the practical applicability of image DDPMs.


For a model trained with $T$ diffusion steps, we would typically sample using the same sequence of $t$ values $(1, 2, ..., T)$ as used during training. However, it is also possible to sample using an arbitrary subsequence $S$ of $t$ values. Given the training noise schedule $\bar{\alpha}_t$, for a given sequence $S$ we can obtain the sampling noise schedule $\bar{\alpha}_{S_t}$, which can be then used to obtain corresponding sampling variances
\begin{equation}
\beta_{S_{t}} = 1 - \frac{\bar{\alpha}_{S_t}}{\bar{\alpha}_{S_{t-1}}},\;\;\;\tilde{\beta}_{S_t} = \frac{1-\bar{\alpha}_{S_{t-1}}}{1-\bar{\alpha}_{S_t}} \beta_{S_t}
\end{equation}

Now, since $\Sigma_{\theta}(x_{S_t}, S_t)$ is parameterized as a range between $\beta_{S_t}$ and $\tilde{\beta}_{S_t}$, it will automatically be rescaled for the shorter diffusion process. We can thus compute $p(x_{S_{t-1}}|x_{S_t})$ as $\mathcal{N}(\mu_{\theta}(x_{S_t}, S_t), \Sigma_{\theta}(x_{S_t}, S_t))$.

\begin{figure}[t]
\begin{center}
\centerline{\includegraphics[width=0.8\columnwidth]{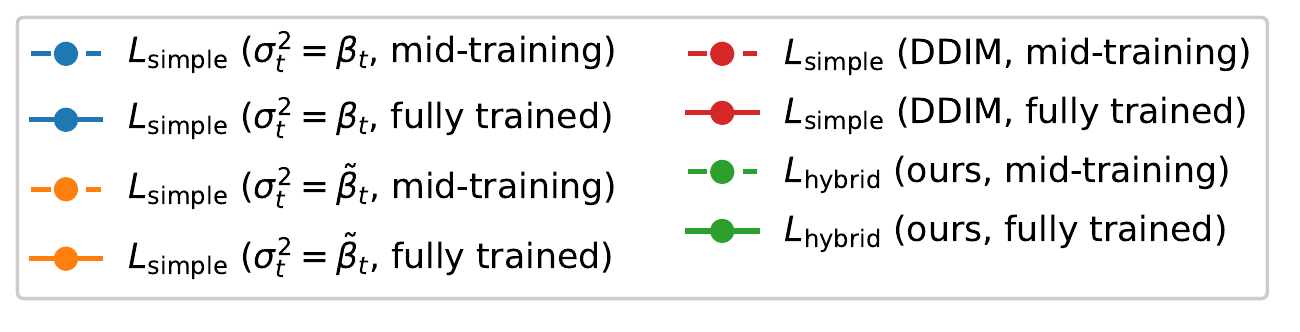}}
\centerline{\includegraphics[width=0.8\columnwidth]{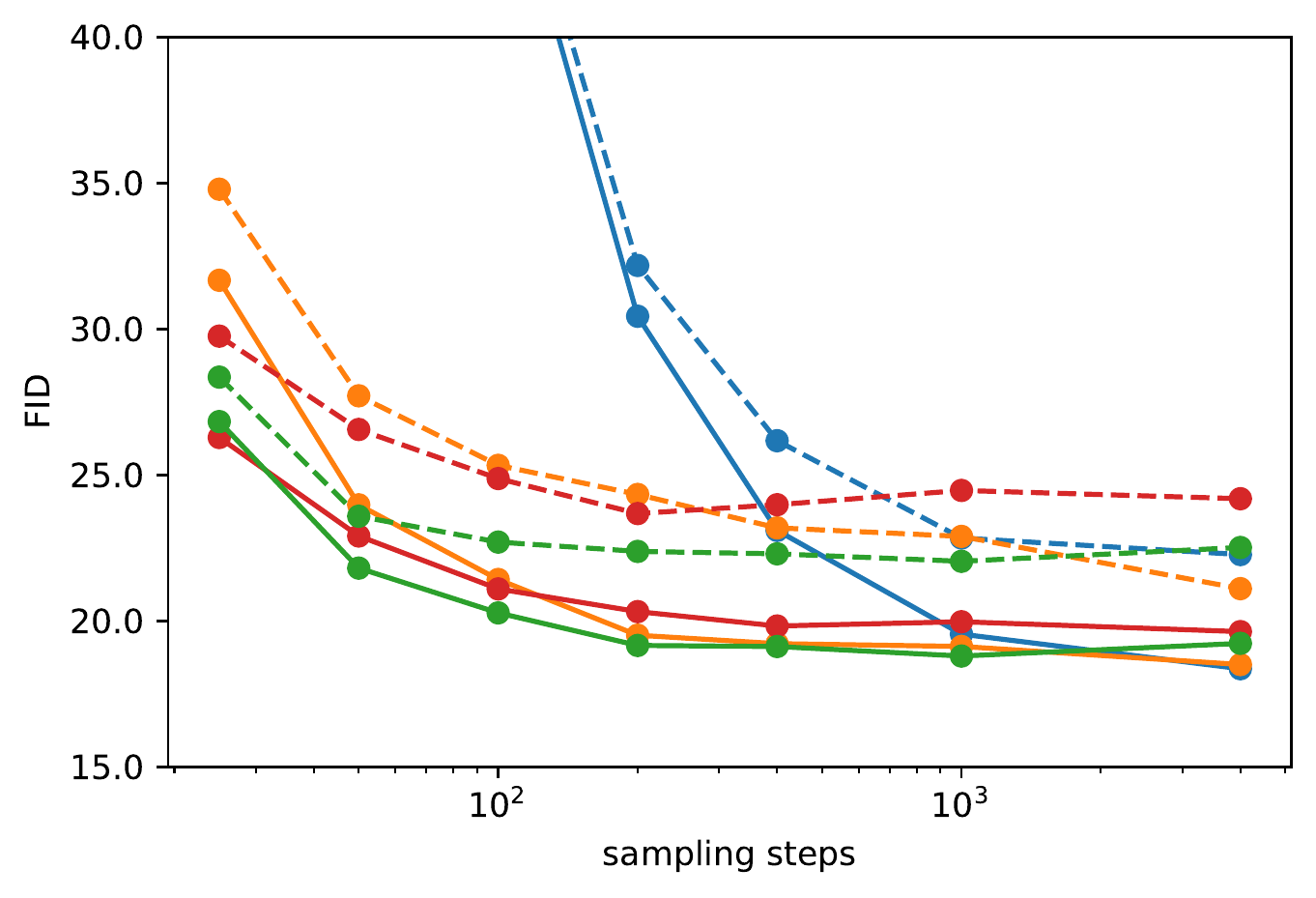}}
\centerline{\includegraphics[width=0.8\columnwidth]{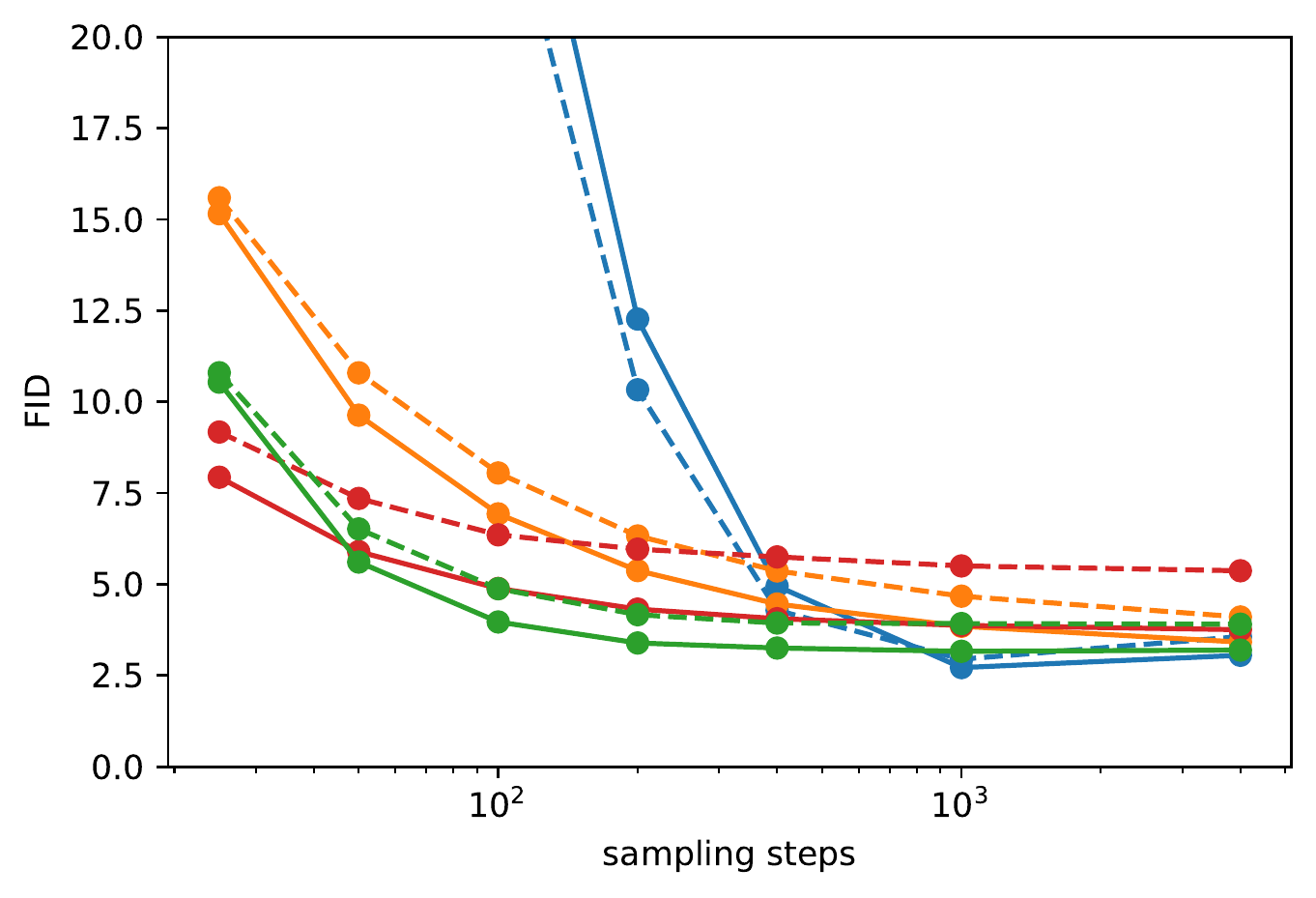}}
\caption{\label{fig:fidvssteps} FID versus number of sampling steps, for models trained on ImageNet $64 \times 64$ (top) and CIFAR-10 (bottom). All models were trained with 4000 diffusion steps.}
\end{center}
\vskip -0.4in
\end{figure}

To reduce the number of sampling steps from $T$ to $K$, we use $K$ evenly spaced real numbers between 1 and $T$ (inclusive), and then round each resulting number to the nearest integer. In Figure \ref{fig:fidvssteps}, we evaluate FIDs for an $L_{\text{hybrid}}$ model and an $L_{\text{simple}}$ model that were trained with 4000 diffusion steps, using 25, 50, 100, 200, 400, 1000, and 4000 sampling steps. We do this for both a fully-trained checkpoint, and a checkpoint mid-way through training. For CIFAR-10 we used 200K and 500K training iterations, and for ImageNet-64 we used 500K and 1500K training iterations.
We find that the $L_{\text{simple}}$ models with fixed sigmas (with both the larger $\sigma_t^2 = \beta_t$ and the smaller $\sigma_t^2 = \tilde{\beta}_t$) suffer much more in sample quality when using a reduced number of sampling steps, whereas our $L_{\text{hybrid}}$ model with learnt sigmas maintains high sample quality. With this model, 100 sampling steps is sufficient to achieve near-optimal FIDs for our fully trained models. 

Parallel to our work, \citet{ddim} propose a fast sampling algorithm for DDPMs by producing a new implicit model that has the same marginal noise distributions, but deterministically maps noise to images. We include their algorithm, DDIM, in Figure \ref{fig:fidvssteps}, finding that DDIM produces better samples with fewer than 50 sampling steps, but worse samples when using 50 or more steps. Interestingly, DDIM performs worse at the start of training, but closes the gap to other samplers as training continues. We found that our striding technique drastically reduced performance of DDIM, so our DDIM results instead use the constant striding\footnote{We additionally tried the quadratic stride from \citet{ddim}, but found that it hurt sample quality when combined with our cosine schedule.} from \citet{ddim}, wherein the final timestep is $T-T/K+1$ rather than $T$. The other samplers performed slightly better with our striding.



\section{Comparison to GANs}
\begin{table}[t]
    \caption{\label{tbl:classcondfid} Sample quality comparison on class-conditional ImageNet $64 \times 64$. Precision and recall \citep{improvedpr} are measured using Inception-V3 features and $K=5$. We trained BigGAN-deep for 125K iterations, and did not use truncation for sampling to maximize recall for the GAN.}
    \centering
    \vskip 0.15in
	\begin{center}
	\begin{small}
    \begin{tabular}{lccc}
        \toprule
        Model & FID & Prec. & Recall \\
        \midrule
        BigGAN-deep \citep{biggan} & 4.06 & \bf 0.86 & 0.59 \\
        Improved Diffusion (small) & 6.92 & 0.77 & \bf 0.72 \\ 
        Improved Diffusion (large) & \bf 2.92 & 0.82 & \bf 0.71 \\
        \bottomrule
    \end{tabular}
\end{small}
\end{center}
\vskip -0.1in
\end{table}

\begin{figure}[t]
\begin{center}
\centerline{\includegraphics[width=0.8\columnwidth]{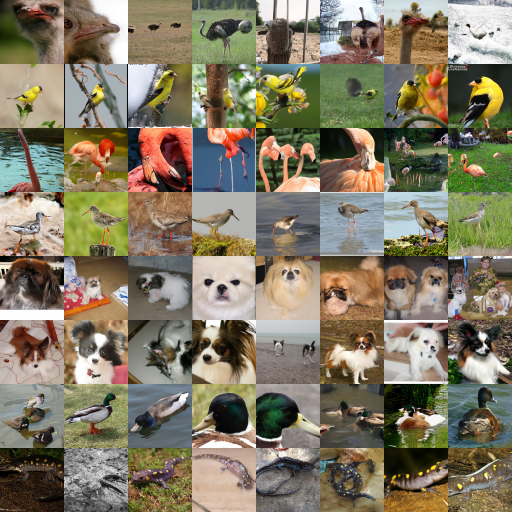}}
\caption{\label{fig:classcond} Class-conditional ImageNet $64 \times 64$ samples generated using 250 sampling steps from $L_{\text{hybrid}}$ model (FID 2.92). The classes are 9: ostrich, 11: goldfinch, 130: flamingo, 141: redshank, 154: pekinese, 157: papillon, 97: drake and 28: spotted salamander. We see that there is a high diversity in each class, suggesting good coverage of the target distribution}
\end{center}
\vskip -0.2in
\end{figure}

While likelihood is a good proxy for mode-coverage, it is difficult to compare to GANs with this metric. Instead, we turn to precision and recall \citep{improvedpr}. Since it is common in the GAN literature to train class-conditional models, we do the same for this experiment. To make our models class-conditional, we inject class information through the same pathway as the timestep $t$. In particular, we add a class embedding $v_i$ to the timestep embedding $e_t$, and pass this embedding to residual blocks throughout the model. We train using the $L_{\text{hybrid}}$ objective and use $250$ sampling steps. We train two models: a "small" model with 100M parameters for 1.7M training steps, and a larger model with 270 million parameters for 250K iterations. We train one BigGAN-deep model with 100M parameters across the generator and discriminator.

When computing metrics for this task, we generated 50K samples (rather than 10K) to be directly comparable to other works.\footnote{We found that using more samples led to a decrease in estimated FID of roughly 2 points.} This is the only ImageNet $64 \times 64$ FID we report that was computed using 50K samples. For FID, the reference distribution features were computed over the full training set, following \cite{biggan}.

Figure \ref{fig:classcond} shows our samples from the larger model, and Table \ref{tbl:classcondfid} summarizes our results. We find that BigGAN-deep outperforms our smaller model in terms of FID, but struggles in terms of recall. This suggests that diffusion models are better at covering the modes of the distribution than comparable GANs.

\section{Scaling Model Size}
\label{sec:scaling}

In the previous sections, we showed algorithmic changes that improved log-likelihood and FID without changing the amount of training compute. However, a trend in modern machine learning is that larger models and more training time tend to improve model performance \citep{scalinglaws,igpt,gpt3}. Given this observation, we investigate how FID and NLL scale as a function of training compute. Our results, while preliminary, suggest that DDPMs improve in a predictable way as training compute increases.

To measure how performance scales with training compute, we train four different models on ImageNet $64 \times 64$ with the $L_{\text{hybrid}}$ objective described in Section \ref{sec:learnsigma}. To change model capacity, we apply a depth multiplier across all layers, such that the first layer has either 64, 96, 128, or 192 channels. Note that our previous experiments used 128 channels in the first layer. Since the depth of each layer affects the scale of the initial weights, we scale the Adam \citep{adam} learning rate for each model by $1/\sqrt{\text{channel multiplier}}$, such that the 128 channel model has a learning rate of 0.0001 (as in our other experiments).

\begin{figure}[t]
\begin{center}
\centerline{\includegraphics[width=0.8\columnwidth]{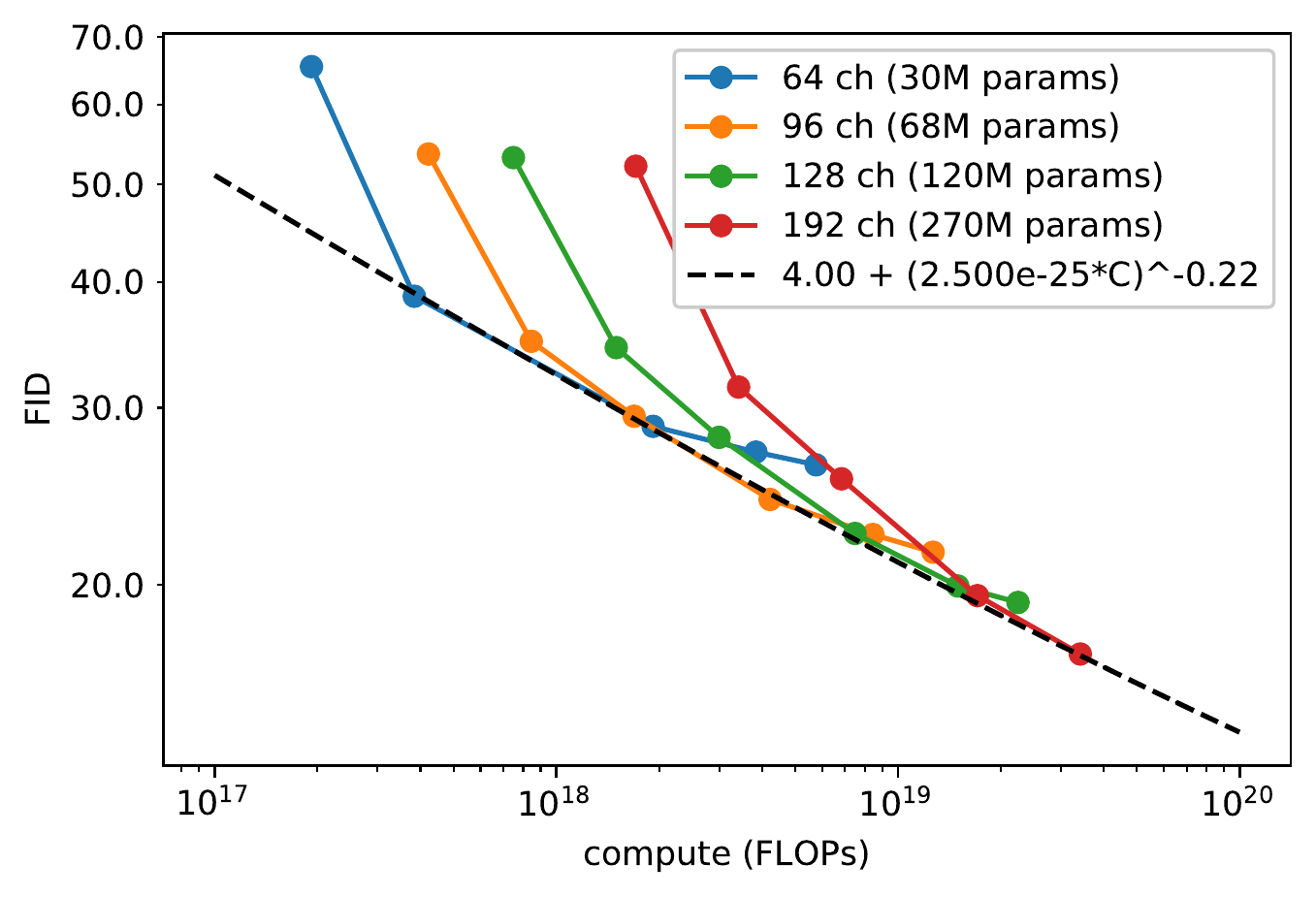}}
\centerline{\includegraphics[width=0.8\columnwidth]{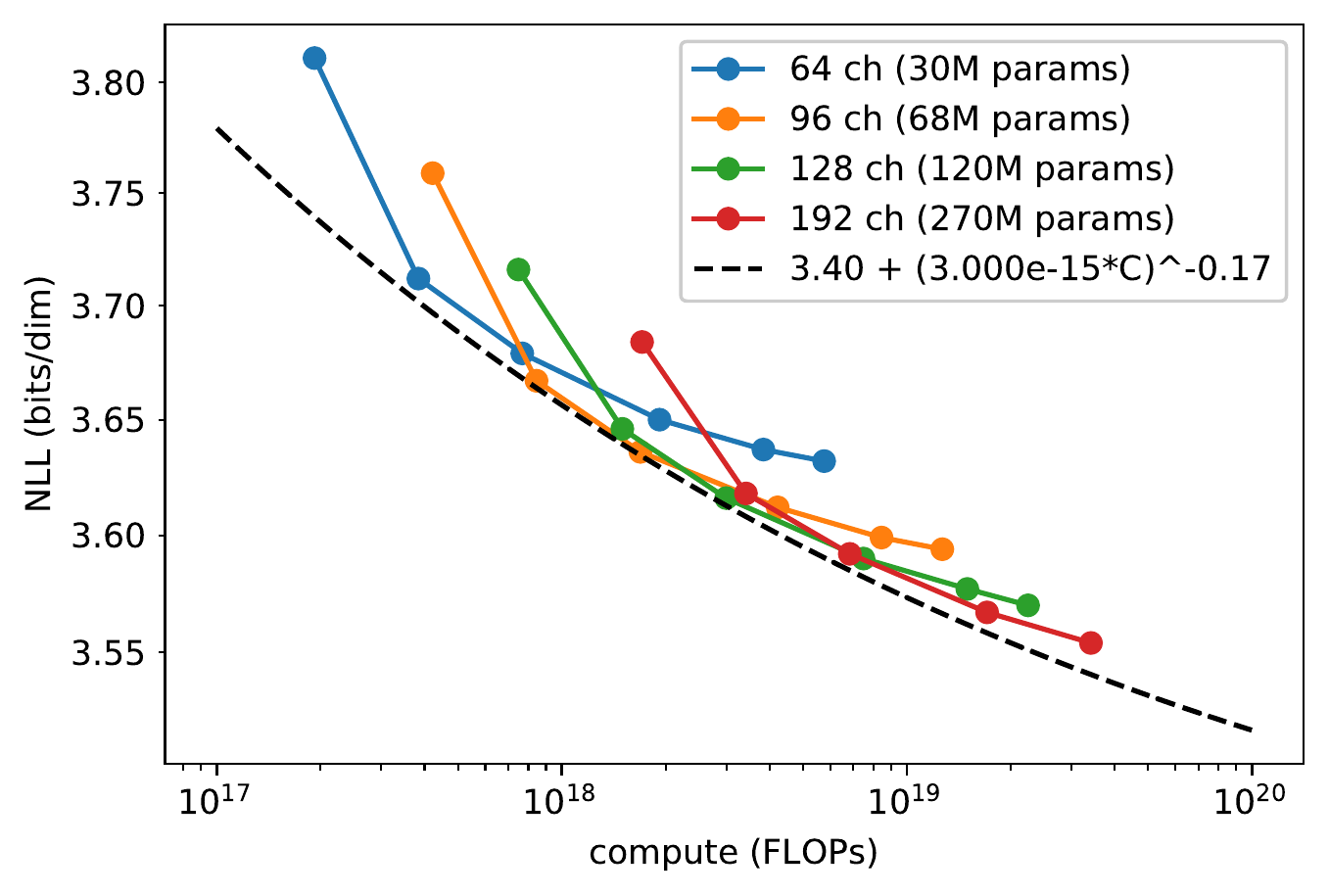}}
\caption{\label{fig:computevsfidnll} FID and validation NLL throughout training on ImageNet $64 \times 64$ for different model sizes. The constant for the FID trend line was approximated using the FID of in-distribution data. For the NLL trend line, the constant was approximated by rounding down the current state-of-the-art NLL \citep{routingtransformer} on this dataset.}
\end{center}
\vskip -0.4in
\end{figure}

Figure \ref{fig:computevsfidnll} shows how FID and NLL improve relative to theoretical training compute.\footnote{The x-axis assumes full hardware utilization} The FID curve looks approximately linear on a log-log plot, suggesting that FID scales according to a power law (plotted as the black dashed line). The NLL curve does not fit a power law as cleanly, suggesting that validation NLL scales in a less-favorable manner than FID. This could be caused by a variety of factors, such as 1) an unexpectedly high irreducible loss \citep{scalingcompendium} for this type of diffusion model, or 2) the model overfitting to the training distribution. We also note that these models do not achieve optimal log-likelihoods in general because they were trained with our $L_{\text{hybrid}}$ objective and not directly with $L_{\text{vlb}}$ to keep both good log-likelihoods and sample quality.
\section{Related Work}
\citet{wavegrad} and \citet{diffwave} are two recent works that use DDPMs to produce high fidelity audio conditioned on mel-spectrograms. Concurrent to our work, \citet{wavegrad} use a combination of improved schedule and $L_1$ loss to allow sampling with fewer steps with very little reduction in sample quality. However, compared to our unconditional image generation task, their generative task has a strong input conditioning signal provided by the mel-spectrograms, and we hypothesize that this makes it easier to sample with fewer diffusion steps.

\citet{adversarial} explored score matching in the image domain, and constructed an adversarial training objective to produce better $x_0$ predictions. However, they found that choosing a better network architecture removed the need for this adversarial objective, suggesting that the adversarial objective is not necessary for powerful generative modeling.

Parallel to our work, \citet{ddim} and \citet{sde} propose fast sampling algorithms for models trained with the DDPM objective by leveraging different sampling processes. \citet{ddim} does this by deriving an implicit generative model that has the same marginal noise distributions as DDPMs while deterministically mapping noise to images. \citet{sde} model the diffusion process as the discretization of a continuous SDE, and observe that there exists an ODE that corresponds to sampling from the reverse SDE. By varying the numerical precision of an ODE solver, they can sample with fewer function evaluations. However, they note that this technique obtains worse samples than ancestral sampling when used directly, and only achieves better FID when combined with Langevin corrector steps. This in turn requires hand-tuning of a signal-to-noise ratio for the Langevin steps. Our method allows fast sampling directly from the ancestral process, which removes the need for extra hyperparameters.

Also in parallel, \citet{ebmdiffusion} develops a diffusion model with reverse diffusion steps modeled by an energy-based model. A potential implication of this approach is that fewer diffusion steps should be needed to achieve good samples.

\section{Conclusion}
We have shown that, with a few modifications, DDPMs can sample much faster and achieve better log-likelihoods with little impact on sample quality. The likelihood is improved by learning $\Sigma_{\theta}$ using our parameterization and $L_{\text{hybrid}}$ objective. This brings the likelihood of these models much closer to other likelihood-based models. We surprisingly discover that this change also allows sampling from these models with many fewer steps. 

We have also found that DDPMs can match the sample quality of GANs while achieving much better mode coverage as measured by recall. Furthermore, we have investigated how DDPMs scale with the amount of available training compute, and found that more training compute trivially leads to better sample quality and log-likelihood.

The combination of these results makes DDPMs an attractive choice for generative modeling, since they combine good log-likelihoods, high-quality samples, and reasonably fast sampling with a well-grounded, stationary training objective that scales easily with training compute. These results indicate that DDPMs are a promising direction for future research.

\bibliography{main}
\bibliographystyle{icml2021}

\clearpage

\appendix

\section{Hyperparameters}
\label{app:hyperparameters}

For all of our experiments, we use a UNet model architecture\footnote{In initial experiments, we found that a ResNet-style architecture with no downsampling achieved better log-likelihoods but worse FIDs than the UNet architecture.} similar to that used by \citet{ddpm}. We changed the attention layers to use multi-head attention \citep{transformers}, and opted to use four attention heads rather than one (while keeping the same total number of channels). We employ attention not only at the 16x16 resolution, but also at the 8x8 resolution. Additionally, we changed the way the model conditions on $t$. In particular, instead of computing a conditioning vector $v$ and injecting it into hidden state $h$ as $\text{GroupNorm}(h + v)$, we compute conditioning vectors $w$ and $b$ and inject them into the hidden state as $\text{GroupNorm}(h)(w + 1) + b$. We found in preliminary experiments on ImageNet $64 \times 64$ that these modifications slightly improved FID.

For ImageNet $64 \times 64$ the architecture we use is described as follows. The downsampling stack performs four steps of downsampling, each with three residual blocks \citep{resnet}. The upsampling stack is setup as a mirror image of the downsampling stack. From highest to lowest resolution, the UNet stages use $[C, 2C, 3C, 4C]$ channels, respectively. In our ImageNet $64 \times 64$ ablations, we set $C=128$, but we experiment with scaling $C$ in a later section. We estimate that, with $C=128$, our model is comprised of 120M parameters and requires roughly 39 billion FLOPs in the forward pass.

For our CIFAR-10 experiments, we use a smaller model with three resblocks per downsampling stage and layer widths $[C, 2C, 2C, 2C]$ with $C=128$. We swept over dropout values $\{0.1, 0.2, 0.3\}$ and found that 0.1 worked best for the linear schedule while 0.3 worked best for our cosine schedule. We expand upon this in Section \ref{app:overfitting}.

We use Adam \citep{adam} for all of our experiments. For most experiments, we use a batch size of 128, a learning rate of $10^{-4}$, and an exponential moving average (EMA) over model parameters with a rate of 0.9999. For our scaling experiments, we vary the learning rate to accomodate for different model sizes. For our larger class-conditional ImageNet $64 \times 64$ experiments, we scaled up the batch size to 2048 for faster training on more GPUs.

When using the linear noise schedule from \citet{ddpm}, we linearly interpolate from $\beta_1=0.0001/4$ to $\beta_{4000}=0.02/4$ to preserve the shape of $\bar{\alpha}_t$ for the $T=4000$ schedule.

When computing FID we produce 50K samples from our models, except for unconditional ImageNet $64 \times 64$ where we produce 10K samples. Using only 10K samples biases the FID to be higher, but requires much less compute for sampling and helps do large ablations. Since we mainly use FID for relative comparisons on unconditional ImageNet $64 \times 64$, this bias is acceptable. For computing the reference distribution statistics we follow prior work \citep{ddpm,biggan} and use the full training set for CIFAR-10 and ImageNet, and 50K training samples for LSUN. Note that unconditional ImageNet $64 \times 64$ models are trained and evaluated using the official ImageNet-64 dataset \cite{pixelrnn}, whereas for class conditional ImageNet $64 \times 64$ and $256 \times 256$ we center crop and area downsample images \cite{biggan}.


\section{Fast Sampling on LSUN $256 \times 256$}
\begin{figure}[ht]
    \centering
    \includegraphics[width=\columnwidth]{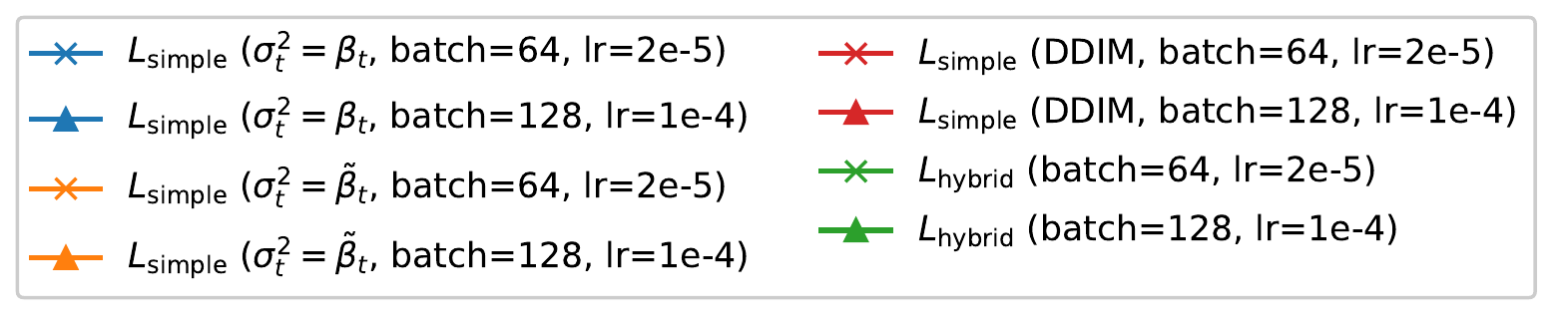}
    \includegraphics[width=\columnwidth]{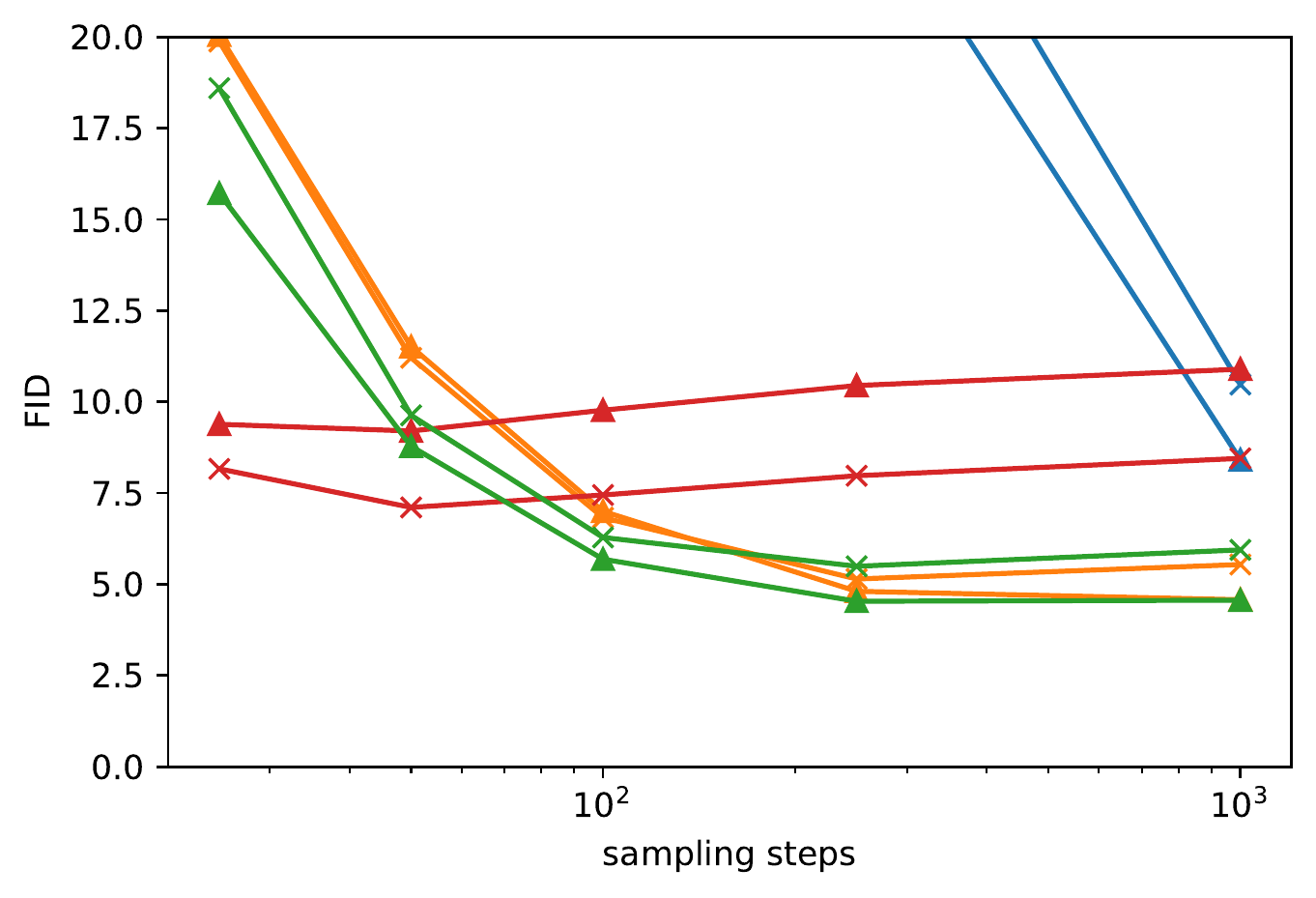}
    \vskip -0.1in
    \caption{\label{fig:fidvssteps_lsun} FID vs. number of sampling steps from an LSUN $256 \times 256$ bedroom model.}
\end{figure}
To test the effectiveness of our $L_{\text{hybrid}}$ models on a high-resolution domain, we trained both $L_{\text{hybrid}}$ and $L_{\text{simple}}$ models on the LSUN bedroom \citep{lsun} dataset. We train two models: one with batch size 64 and learning rate $2 \times 10^{-5}$ as in \citet{ddpm}, and another with a larger batch size 128 and learning rate $10^{-4}$. All models were trained with 153.6M examples, which is 2.4M training iterations with batch size 64.

Our results are displayed in Figure \ref{fig:fidvssteps_lsun}. We find that DDIM outperforms our $L_{\text{hybrid}}$ model when using fewer than 50 diffusion steps, while our $L_{\text{hybrid}}$ model outperforms DDIM with more than 50 diffusion steps. Interestingly, we note that DDIM benefits from a smaller learning rate and batch size, whereas our method is able to take advantage of a larger learning rate and batch size.

\section{Sample Quality on ImageNet $256 \times 256$}
We trained two models on class conditional ImageNet $256 \times 256$. The first is a usual diffusion model that directly models the $256 \times 256$ images. The second model reduces compute by chaining a pretrained $64 \times 64$ model $p(x_{64}|y)$ with another upsampling diffusion model $p(x_{256}|x_{64}, y)$ to upsample images to $256 \times 256$. For the upsampling model, the downsampled image $x_{64}$  is passed as extra conditioning input to the UNet. This is similar to VQ-VAE-2 \citep{vqvae2}, which uses two stages of priors at different latent resolutions to more efficiently learn global and local features. The linear schedule worked better for $256 \times 256$ images, so we used that for these results. Table \ref{tbl:classcond256fid} summarizes our results. For VQ-VAE-2, we use the FIDs reported in \cite{cas}. Diffusion models still obtain the best FIDs for a likelihood-based model, and close the gap to GANs considerably. 

\begin{table}[ht]
    \centering
    \begin{tabular}{l|c}
        \bf MODEL & \bf FID \\
        \hline
        VQ-VAE-2 (\cite{vqvae2}, two-stage) & 38.1 \\
        Improved Diffusion (ours, single-stage) & 31.5 \\
        Improved Diffusion (ours, two-stage) & \bf{12.3} \\ 
        BigGAN \citep{biggan} & 7.7 \\
        BigGAN-deep \citep{biggan} & \bf{7.0} \\
    \end{tabular}
    \caption{\label{tbl:classcond256fid} Sample quality comparison on class conditional ImageNet $256 \times 256$. BigGAN FIDs are reported for the truncation that results in the best FID.}
\end{table}

\begin{figure}[ht]
    \centerline{\includegraphics[width=\columnwidth]{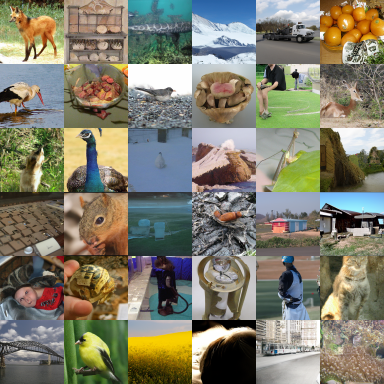}}
    \vskip 0.1in
    \centerline{\includegraphics[width=\columnwidth]{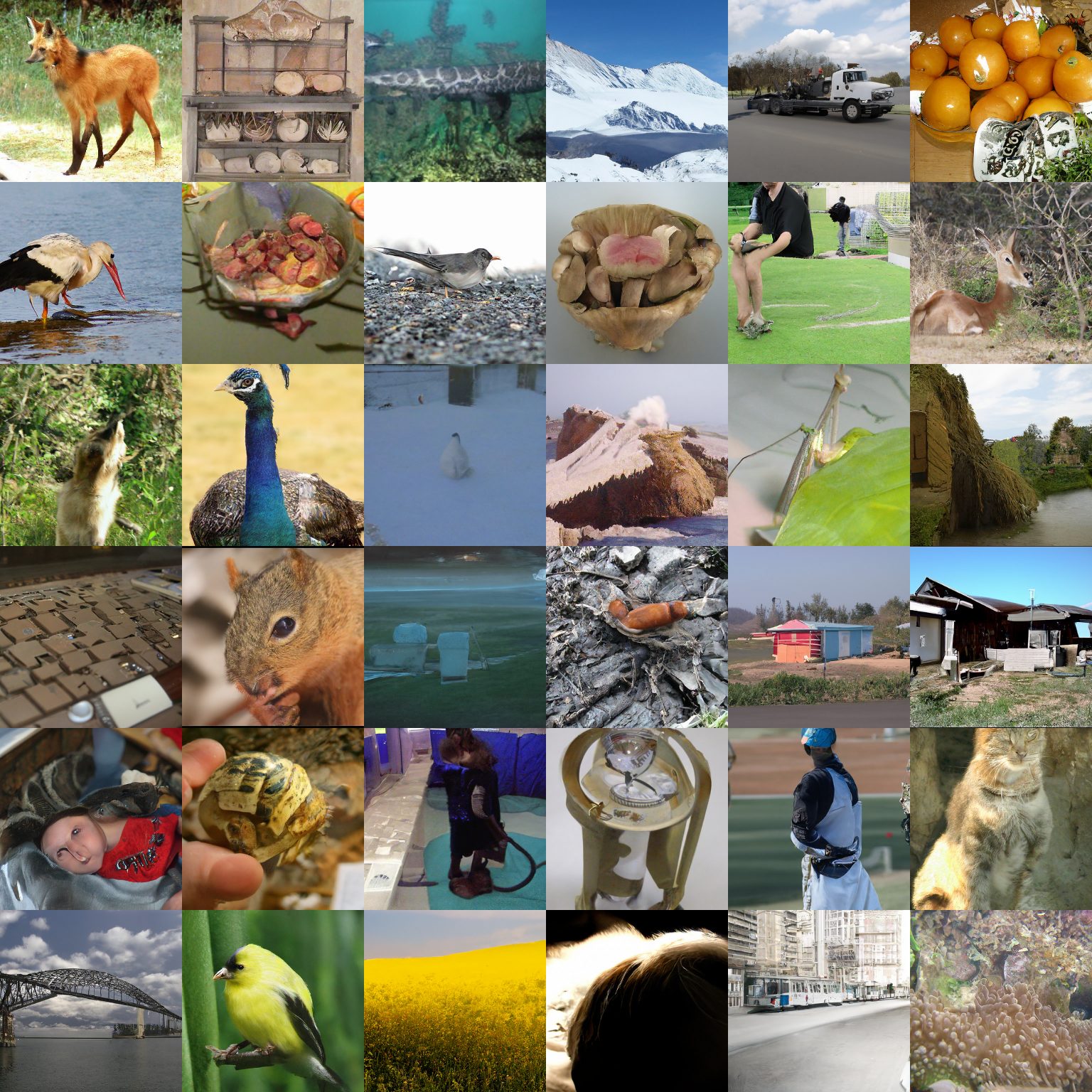}}
    \caption{\label{fig:classcondtwostage}Random samples from two-stage class conditional ImageNet $256 \times 256$ model. On top are random samples from the $64 \times 64$ model (FID 2.92), whereas on bottom are the results after upsampling them to $256 \times 256$ (FID 12.3). Each model uses 250 sampling steps.}
    \vspace{1.5in}
\end{figure}

\clearpage

\section{Combining $L_{\text{hybrid}}$ and $L_{\text{vlb}}$ Models}

\begin{figure}[ht]
    \centerline{\includegraphics[width=\columnwidth]{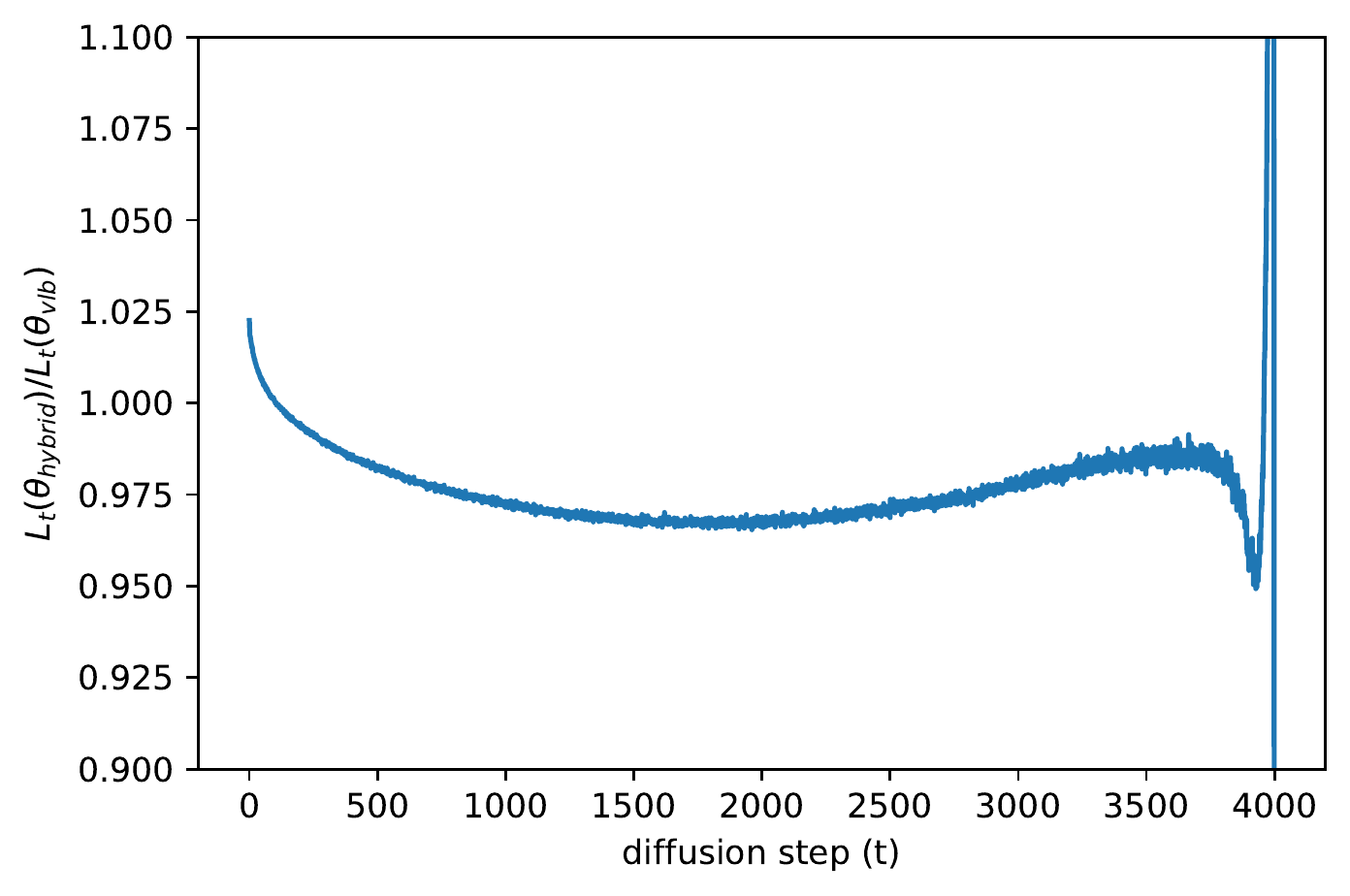}}
    \caption{\label{fig:bpdterms} The ratio between VLB terms for each diffusion step of $\theta_{\text{hybrid}}$ and $\theta_{\text{vlb}}$. Values less than 1.0 indicate that $\theta_{\text{hybrid}}$ is "better" than $\theta_{\text{vlb}}$ for that timestep of the diffusion process.}
\end{figure}

\begin{figure}[ht]
    \centering
    \includegraphics[width=\columnwidth]{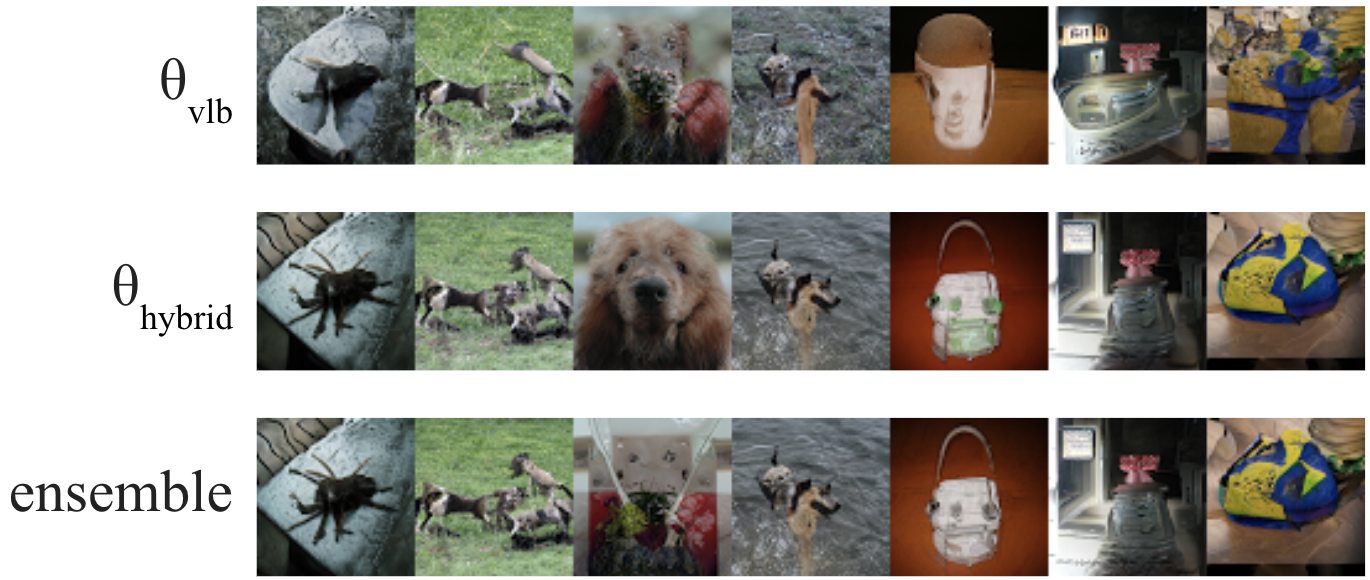}
    \caption{\label{fig:objectivesamples} Samples from $\theta_{\text{vlb}}$ and $\theta_{\text{hybrid}}$, as well as an ensemble produced by using $\theta_{\text{vlb}}$ for the first and last 100 diffusion steps. For these samples, the seed was fixed, allowing a direct comparison between models.} 
\end{figure}

To understand the trade-off between $L_{\text{hybrid}}$ and $L_{\text{vlb}}$, we show in Figure \ref{fig:bpdterms} that the model resulting from $L_{\text{vlb}}$ (referred to as $\theta_{\text{vlb}}$) is better at the start and end of the diffusion process, while the model resulting from $L_{\text{hybrid}}$ (referred to as $\theta_{\text{hybrid}}$) is better throughout the middle of the diffusion process. This suggests that $\theta_{\text{vlb}}$ is focusing more on imperceptible details, hence the lower sample quality.

Given the above observation, we performed an experiment on ImageNet $64 \times 64$ to combine the two models by constructing an ensemble that uses $\theta_{\text{hybrid}}$ for $t \in [100, T-100)$ and $\theta_{\text{vlb}}$ elsewhere. We found that this model achieved an FID of \textbf{19.9} and an NLL of \textbf{3.52 bits/dim}. This is only slightly worse than $\theta_{\text{hybrid}}$ in terms of FID, while being better than both models in terms of NLL.

\newpage
\section{Log-likelihood with Fewer Diffusion Steps}

\begin{figure}[ht]
    \centering
    \includegraphics[width=0.85\columnwidth]{fid_vs_steps_cifar_legend-eps.pdf}
    \includegraphics[width=\columnwidth]{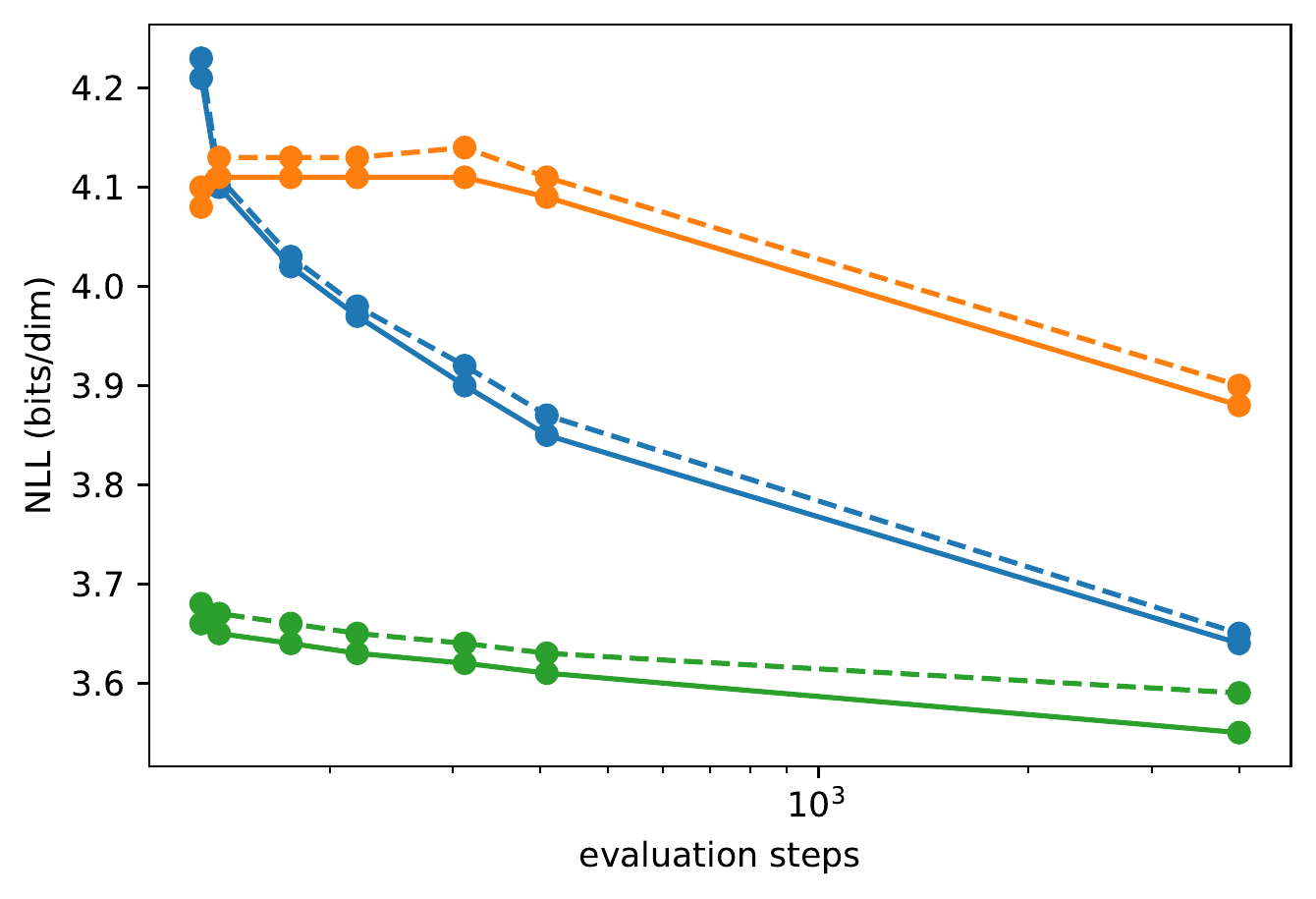}
    \includegraphics[width=\columnwidth]{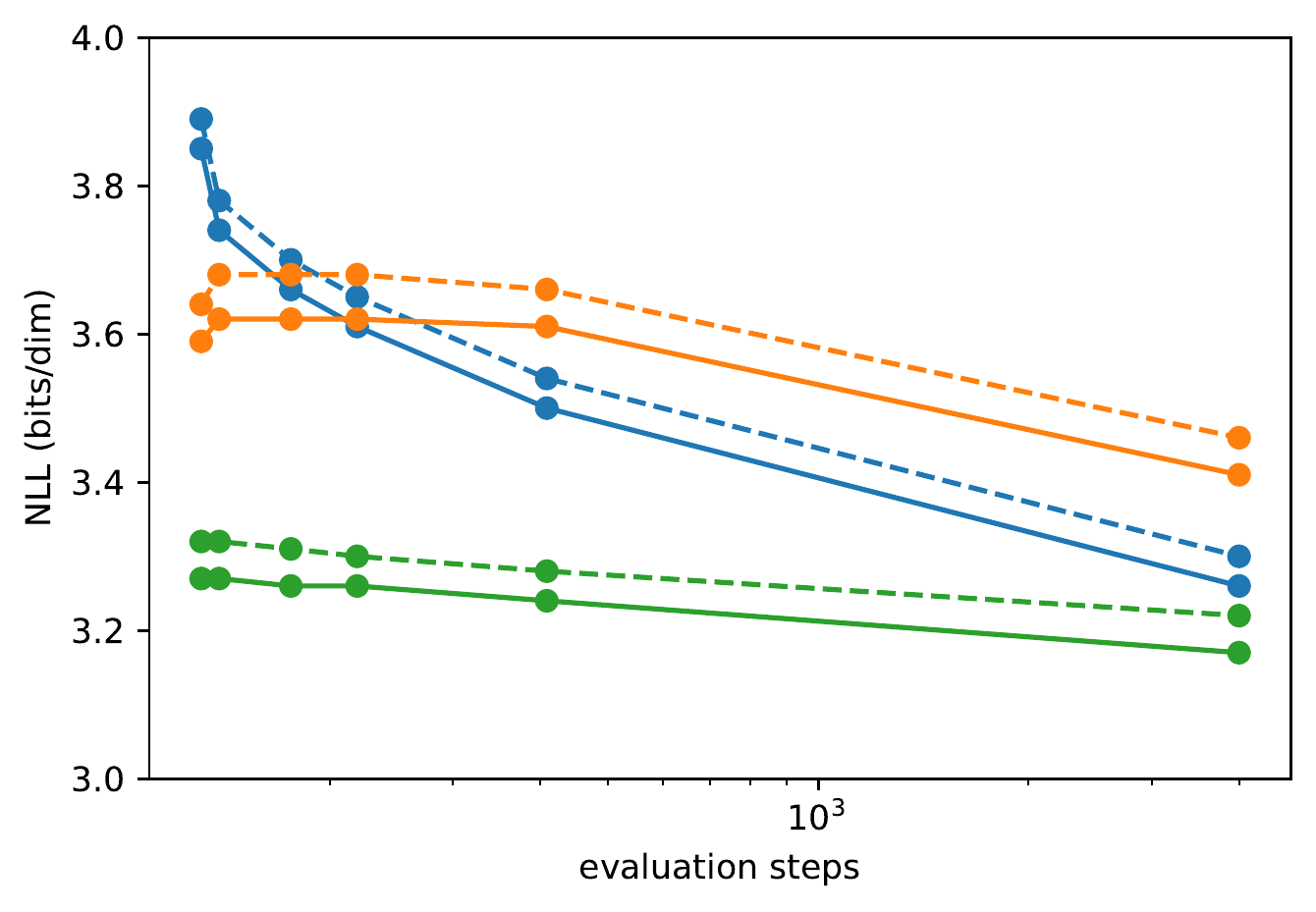}
    \caption{\label{fig:nllvssteps} NLL versus number of evaluation steps, for models trained on ImageNet $64 \times 64$ (top) and CIFAR-10 (bottom). All models were trained with 4000 diffusion steps.}
\end{figure}

Figures \ref{fig:nllvssteps} plots negative log-likelihood as a function of number of sampling steps for both ImageNet $64 \times 64$ and CIFAR-10. In initial experiments, we found that although constant striding did not significantly affect FID, it drastically reduced log-likelihood. To address this, we use a strided subset of timesteps as for FID, but we also include every $t$ from 1 to $T/K$. This requires $T/K$ extra evaluation steps, but greatly improves log-likelihood compared to the uniformly strided schedule. We did not attempt to calculate NLL using DDIM, since \citet{ddim} does not present NLL results or a simple way of estimating likelihood under DDIM.

\clearpage

\section{Overfitting on CIFAR-10}
\label{app:overfitting}

\begin{figure}[ht]
    \centering
    \includegraphics[width=\columnwidth]{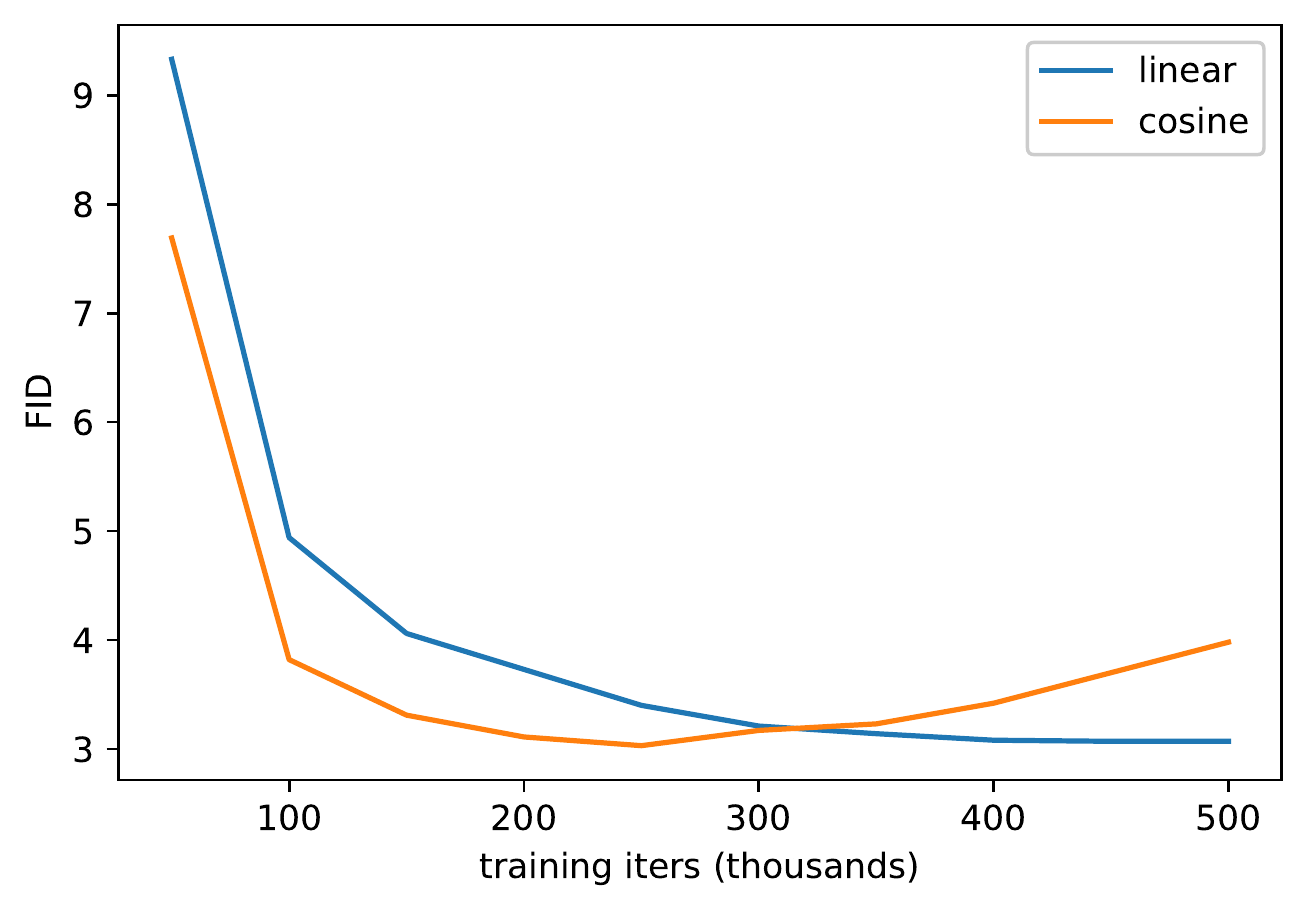}
    \includegraphics[width=\columnwidth]{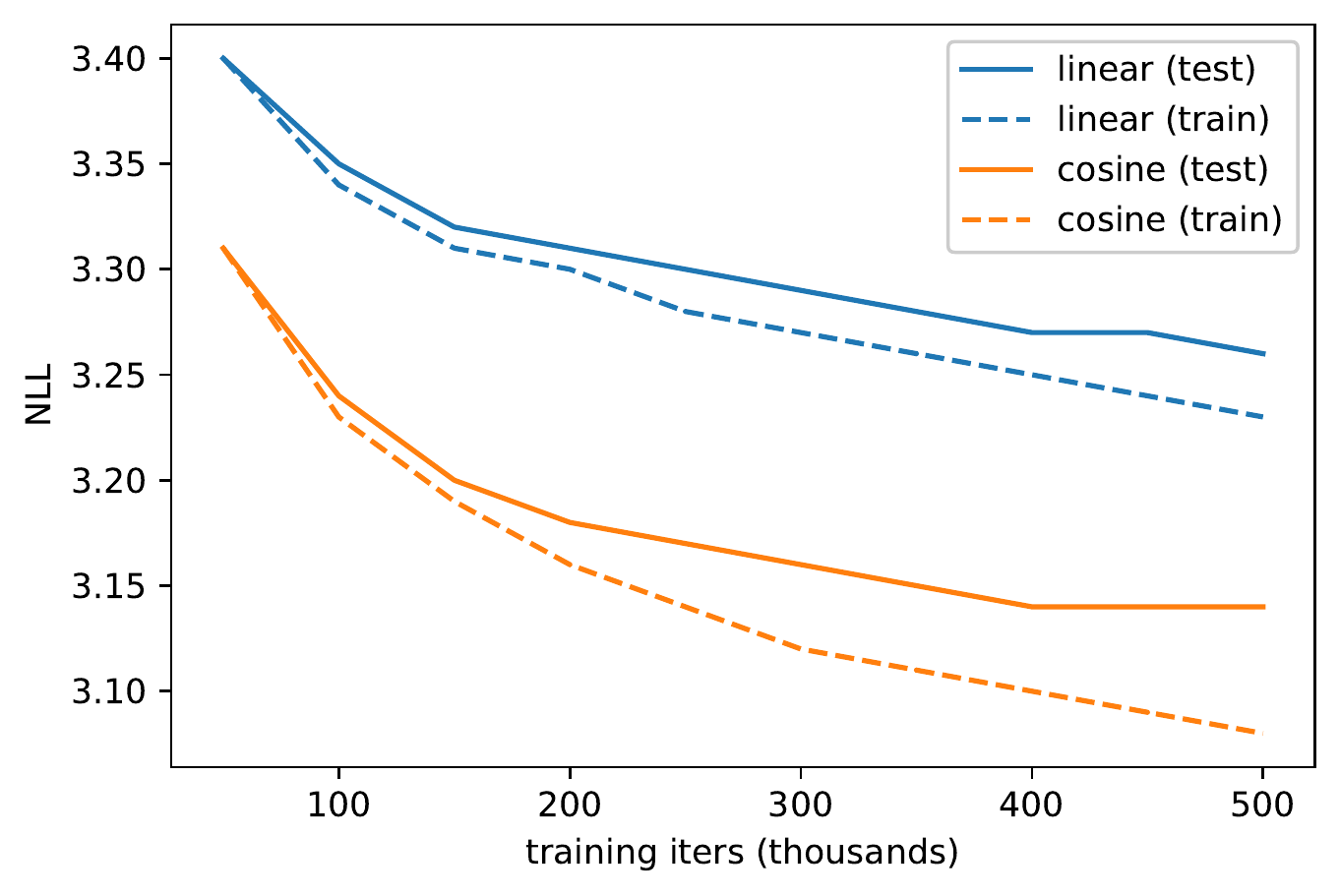}
    \caption{\label{fig:overfitting} FID (top) and NLL (bottom) over the course of training for two CIFAR-10 models, both with dropout 0.1. The model trained with the linear schedule learns more slowly, but does not overfit as quickly. When too much overfitting occurs, we observed overfitting artifacts similar to those from \citet{pixelcnn++}, which is reflected by increasing FID.}
\end{figure}

On CIFAR-10, we noticed that all models overfit, but tended to reach similar optimal FID at some point during training. Holding dropout constant, we found that models trained with our cosine schedule tended to reach optimal performance (and then overfit) more quickly than those trained with the linear schedule (Figure \ref{fig:overfitting}). In our experiments, we corrected for this difference by using more dropout for our cosine models than the linear models. We suspect that the overfitting from the cosine schedule is either due to 1) less noise in the cosine schedule providing less regularization, or 2) the cosine schedule making optimization, and thus overfitting, easier.

\newpage
\section{Early stopping for FID}
\begin{figure}[ht]
    \centering
    \includegraphics[width=\columnwidth]{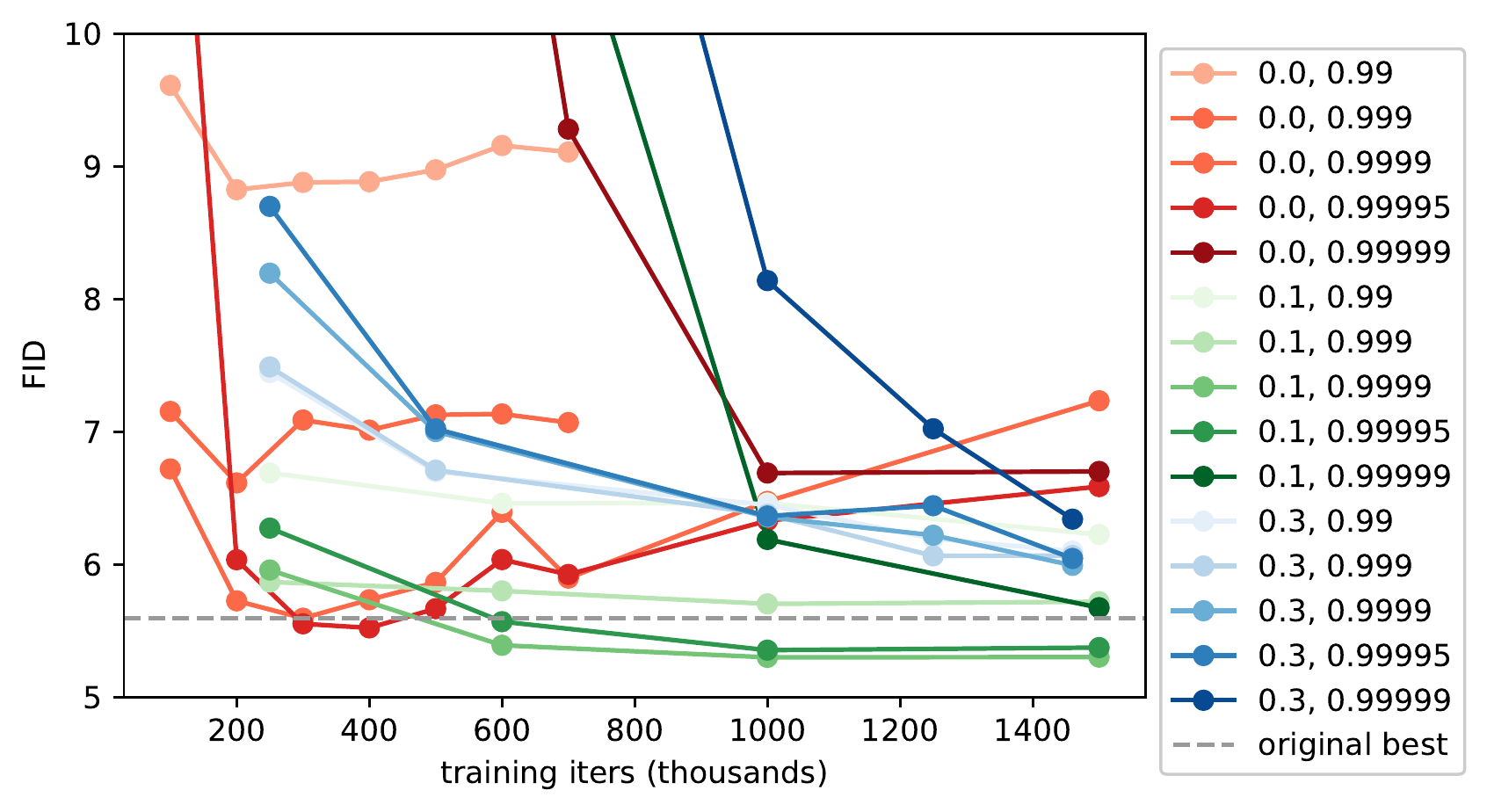}
    \caption{\label{fig:emasweep} A sweep of dropout and EMA hyperparameters on class conditional ImageNet-64.}
\end{figure}
Like on CIFAR-10, we surprisingly observed overfitting on class-conditional ImageNet $64 \times 64$, despite it being a much larger and more diverse dataset. The main observable result of this overfitting was that FID started becoming worse over the course of training. We initially tried a sweep (Figure \ref{fig:emasweep}) over the EMA hyperparameter to make sure it was well tuned, and found that 0.9999 and 0.99995 worked best. We then tried runs with dropout 0.1 and 0.3, and found that models with a small amount of dropout improved the best attainable FID but took longer to get to the same performance and still eventually overfit. We concluded that the best way to train, given what we know, is to early stop and instead increase model size if we want to use additional training compute.

\section{Samples with Varying Steps and Objectives}
\label{app:samples}

\begin{figure}[h!]
    \centerline{\includegraphics[width=0.78\columnwidth]{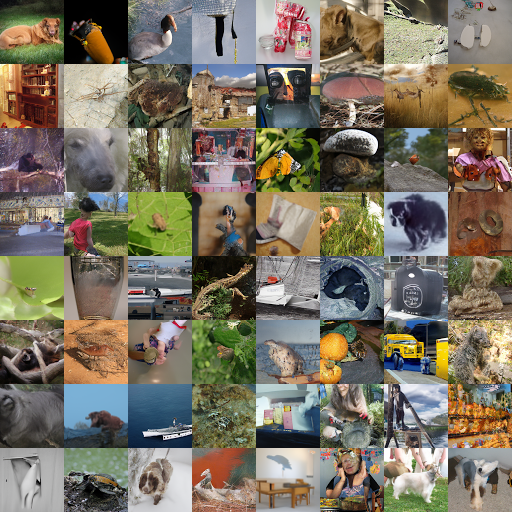}}
    \caption{\label{fig:firstimagenet} 50 sampling steps on unconditional ImageNet $64 \times 64$}
\end{figure}

\begin{figure}[h!]
    \centerline{\includegraphics[width=0.78\columnwidth]{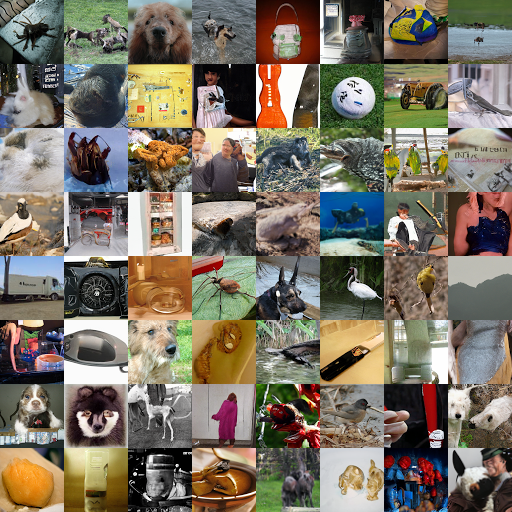}}
    \caption{\label{fig:lastimagenet} 4K sampling steps on unconditional ImageNet $64 \times 64$.}
    \vskip -0.4in
\end{figure}

Figures \ref{fig:firstimagenet} through \ref{fig:lastimagenet} show unconditional ImageNet $64 \times 64$ samples as we reduce number of sampling steps for an $L_{\text{hybrid}}$ model with $4K$ diffusion steps trained for 1.5M training iterations.

\begin{figure}[h!]
    \centerline{\includegraphics[width=0.78\columnwidth]{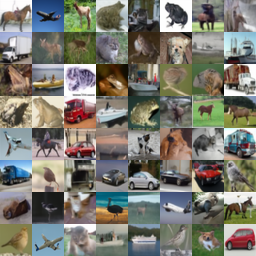}}
    \caption{\label{fig:firstcifar} 50 sampling steps on unconditional CIFAR-10}
\end{figure}

\begin{figure}[h!]
    \centerline{\includegraphics[width=0.78\columnwidth]{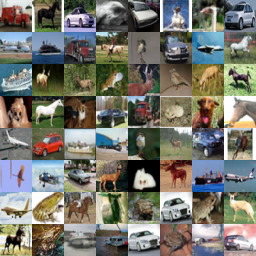}}
    \caption{\label{fig:lastcifar} 4000 sampling steps on unconditional CIFAR-10}
    \vskip -0.4in
\end{figure}

Figures \ref{fig:firstcifar} through \ref{fig:lastcifar} show unconditional CIFAR-10 samples as we reduce number of sampling steps for an $L_{\text{hybrid}}$ model with $4K$ diffusion steps trained for 500K training iterations.

\begin{figure}[t]
    \centerline{\includegraphics[width=0.78\columnwidth]{samples_hybrid_1.5M_grid.png}}
    \vskip 0.1in
    \centerline{\includegraphics[width=0.78\columnwidth]{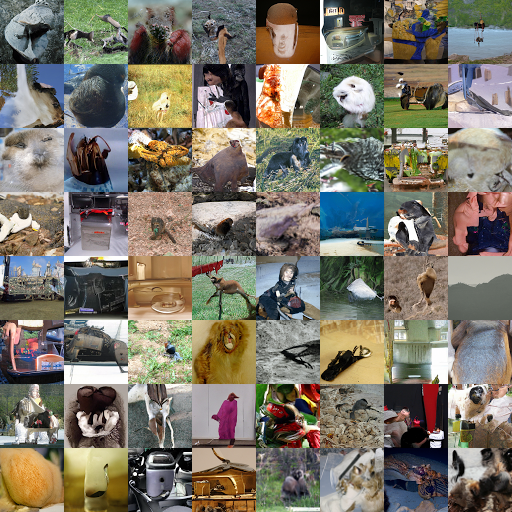}}
    \caption{\label{fig:uncondvlbcomparison} Unconditional ImageNet $64 \times 64$ samples generated from $L_{\text{hybrid}}$ (top) and $L_{\text{vlb}}$ (bottom) models using the exact same random noise. Both models were trained for 1.5M iterations.}
\end{figure}

\begin{figure}[t]
    \centering
    \centerline{\includegraphics[width=0.78\columnwidth]{samples_cifar_hybrid.png}}
    \vskip 0.1in
    \centerline{\includegraphics[width=0.78\columnwidth]{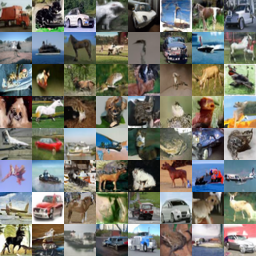}}
    \caption{\label{fig:uncondvlbcomparison_cifar} Unconditional CIFAR-10 samples generated from $L_{\text{hybrid}}$ (top) and $L_{\text{vlb}}$ (bottom) models using the exact same random noise. Both models were trained for 500K iterations.}
\end{figure}

Figures \ref{fig:uncondvlbcomparison} and \ref{fig:uncondvlbcomparison_cifar} highlight the difference in sample quality between models trained with $L_{\text{hybrid}}$ and $L_{\text{vlb}}$.

\clearpage

\begin{figure}[h!]
    \centerline{\includegraphics[width=0.78\columnwidth]{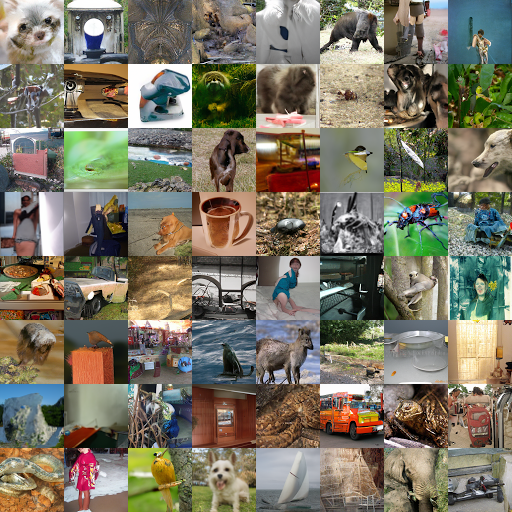}}
    \caption{100 sampling steps on unconditional ImageNet $64 \times 64$}
\end{figure}
\begin{figure}[h!]
    \centerline{\includegraphics[width=0.78\columnwidth]{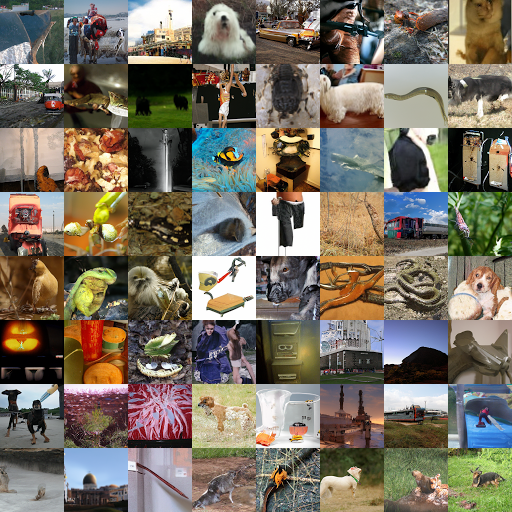}}
    \caption{200 sampling steps on unconditional ImageNet $64 \times 64$}
    \vskip -0.4in
\end{figure}

\newpage
\begin{figure}[h!]
    \centerline{\includegraphics[width=0.78\columnwidth]{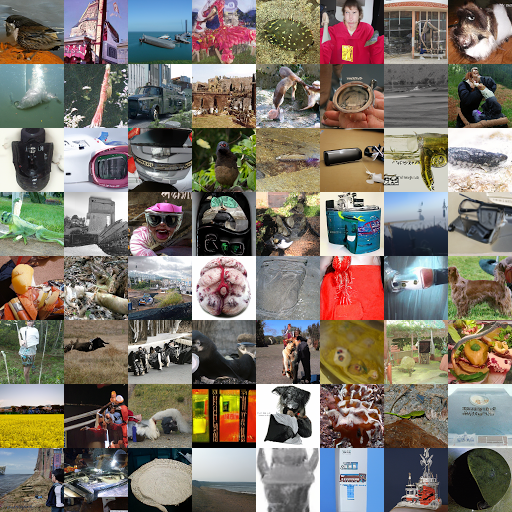}}
    \caption{400 sampling steps on unconditional ImageNet $64 \times 64$}
\end{figure}
\begin{figure}[h!]
    \centerline{\includegraphics[width=0.78\columnwidth]{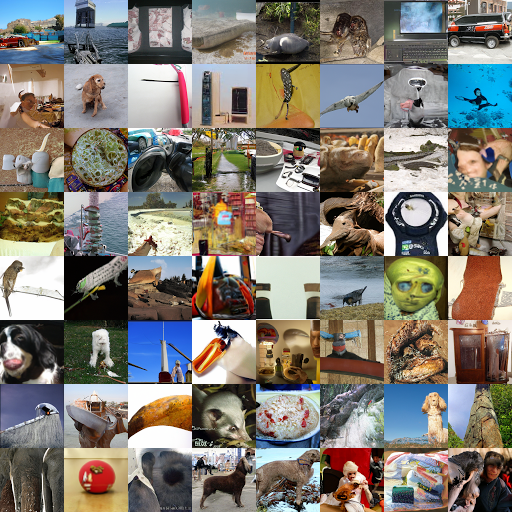}}
    \caption{1000 sampling steps on unconditional ImageNet $64 \times 64$}
\end{figure}

\clearpage

\begin{figure}[h!]
    \centerline{\includegraphics[width=0.78\columnwidth]{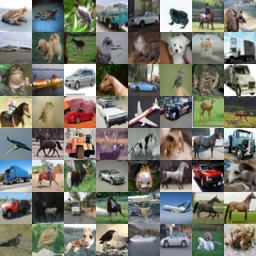}}
    \caption{100 sampling steps on unconditional CIFAR-10}
\end{figure}
\begin{figure}[h!]
    \centerline{\includegraphics[width=0.78\columnwidth]{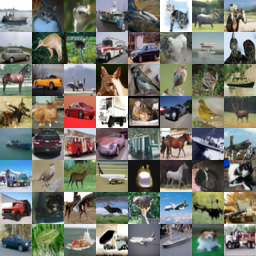}}
    \caption{200 sampling steps on unconditional CIFAR-10}
    \vskip -0.4in
\end{figure}
\newpage
\begin{figure}[h!]
    \centerline{\includegraphics[width=0.78\columnwidth]{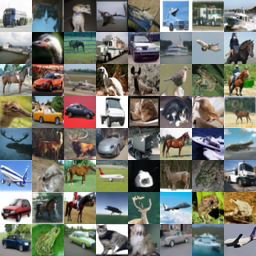}}
    \caption{400 sampling steps on unconditional CIFAR-10}
\end{figure}
\begin{figure}[h!]
    \centerline{\includegraphics[width=0.78\columnwidth]{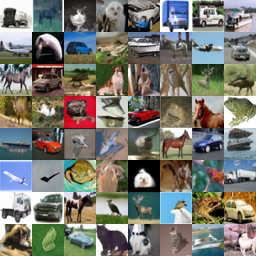}}
    \caption{1000 sampling steps on unconditional CIFAR-10}
\end{figure}

\clearpage

\end{document}